# From Physician Expertise to Clinical Agents: Preserving, Standardizing, and Scaling Physicians' Medical Expertise with Lightweight LLM

Chanyong Luo[3,a], Jirui Dai[5,a], Zhendong Wang[4,a], Kui Chen[10], Jiaxi Yang[9], Bingjie Lu[11], Jing Wang[7], Jiaxin Hao[6], Bing Li[6], Ruiyang He[6], Yiyu Qiao[6], Chenkai Zhang[8], Kaiyu Wang[8], Zhi Liu[2,*], Zeyu Zheng[8,*], Yan Li[6,*], Xiaohong Gu[1,*]

[1] the School of Chinese Medicine, the Beijing University of Chinese Medicine, Beijing, China

[2] the School of Pharmacy, Nanjing University of Chinese Medicine, Nanjing, China

[3] the Infectious disease department, Dongfang Hospital, Beijing University of Chinese Medicine, Beijing, China

[4] the Gulou Hospital of Traditional Chinese Medicine of Beijing, Beijing, China

[5] the Department of Computer Science, Johns Hopkins University, Baltimore, USA

[6] the Department of Education, Dongzhimen Hospital, Beijing University of Chinese Medicine, Beijing, China

[7] the Department of Pediatrics, Wangjing Hospital, China Academy of Chinese Medical Sciences, Beijing, China

[8] the School of Information Engineering, Huzhou University, Huzhou, China

[9] Research Center for Scientific Data Hub, Zhejiang Lab, Hangzhou, China

[10] the Frontier Basic Research Center, Zhejiang Lab, Hangzhou, China

[11] the Research Center for High Efficiency Computing Infrastructure, Zhejiang Lab, Hangzhou, China

## Abstract

Medicine is fundamentally an empirical body of knowledge accumulated through long-term observation, validation, and refinement, and ultimately realized in the messy, high-variance real clinical practice. Physicians' diagnostic-and-therapeutic competence is gradually forged through repeated cycles of "application–reflection–improvement," crystallizing into distinctive, individualized methodologies. Considering the fact of the substantial variation in treatment outcomes, the formation of master physicians' knowledge systems is necessary but time-consuming and their transmission is often limited in scope, making high-quality expertise difficult to scale and disseminate, and the scarcity of advanced clinical resources. To mitigate these challenges, we propose **Med-Shicheng**, a general framework that enables large language model to systematically learn and transfer distinguished physicians' diagnostic-and-therapeutic philosophy, and case-dependent adaptation rules in a standardized manner. Built upon Tianyi, Med-Shicheng comprises five elaborately designed five stages. We take five National Masters of Chinese Medicine/distinguished TCM physicians as representative targets, curate and organize multi-source materials and train a single model to simultaneously internalize the five distinctive knowledge systems across seven tasks: **etiology–pathogenesis analysis**, s**yndrome diagnosis**, **determination of treatment principles**, **prescription generation**, **prescription explanation**, **symptom evolution with corresponding regimen modifications**, and **clinical advice**. Implemented on Qwen2.5-1.5B-

---

[a] The first three authors are the co-first authors.

* Corresponding authors: Zhi Liu (zhiliu@njucm.edu.cn), Xiaohong Gu (642011@bucm.edu.cn), Zeyu Zheng (03243@zju.edu.cn), Yan Li (202250006@bucm.edu.cn)

**Base**, Med-Shicheng can be deployed on resource-constrained GPUs, while demonstrating evaluation performance comparable to **DeepSeek-R1** and **GPT-5**. Besides, we further analyze the reliability of **LLM-as-a-Judge** against physician-based assessment. It shows that although automated judging captures broad performance trends, it exhibits noticeable biases in fine-grained, individualized clinical distinctions, indicating that physician involvement remains necessary when ground truth is unavailable and that judge models require targeted medical domain adaptation to achieve reliable evaluation.

## Keywords

## Introduction

Medicine, in all its forms, is fundamentally an empirical science built upon centuries of accumulated practical knowledge. This holds true whether we speak of Traditional Medicine systems—such as Traditional Chinese Medicine, Mongolian Medicine, Tibetan Medicine, Ayurveda, Unani, or North American Naturopathy—which rely on observing clinical symptoms and macro-phenomenological induction, or Modern Medicine, which is grounded in biochemical markers and macroscopic signal trials and analysis[1,2]. Clinical practice represents a major application of this knowledge system. It is a highly complex endeavor that demands precision and care. One of its defining challenges is patient-specific variability: even among individuals diagnosed with the same disease, symptoms can differ significantly. Unlike in controlled laboratory settings, clinicians must conduct in-depth analyses of each case. They integrate symptoms and etiology to determine the best possible treatment. Mastering this process requires years of hands-on practice. Through experience, doctors learn to apply existing medical knowledge, recognize its limitations, and develop personalized diagnostic and therapeutic methodology tailored to specific diseases and individual patients.

However, disparities in cure rates among clinicians treating identical conditions persist due to heterogeneous factors: differences in experiential knowledge, theoretical mastery, and case analysis proficiency. A physician's cure rate reflects their systematic understanding of a specific disease etiology, patient individuality, and therapeutic strategies—higher rates signal more specialized and precise expertise. Yet, such high-performing clinicians remain scarce, as attaining this level demands prolonged clinical experience accumulation. First, as a prerequisite, clinicians must study a substantial number of professional books, comprehending and mastering their systematic medical knowledge. Second, whether learning from historical cases or engaging in clinical practice, they need extensive exposure to clinical patients, summarizing and reflecting on the medical diagnostic and therapeutic knowledge system during clinical diagnostic-and-therapeutic (D&T). Then, through the iterative process of clinical practice, reflection to cases, and theoretical study, clinicians will develop a personalized theoretical and practical methodology for a specific category of diseases. This methodology includes their analysis of the disease's etiology and pathogenesis, summaries of common symptoms, evaluation of treatment methods, and precautions and contraindications during diagnostic-and-therapeutic. Subsequently, this personalized D&T methodology is continuously refined and improved based on further medical theories, knowledge, and case studies in ongoing clinical practice—a process that typically takes decades.

Beyond the time-intensive nature of expertise formation, such personalized D&T knowledge system dissemination and transmission face inefficiencies and difficulties. For instance, eminent TCM physicians typically pass down expertise via apprenticeship within a small circle, limiting reach to few trainees. This bottleneck exacerbates disparities in frontline clinicians—particularly inexperienced young clinicians' skill development and concentrates high-demand patients (e.g., complex cases) among scarce elite physicians, amplifying healthcare resource inequities. Thus, standardized, scalable methods to archive and transfer individualized expertise hold transformative potential for both clinical practice and medical education. Given the cognitive limits of human physicians, constructing a "second brain" to systematically model elite physicians' diagnostic reasoning, therapeutic principles, and case-specific adaptations in a standardized and scalable way—then disseminate this knowledge efficiently—remains an open and urgent challenge.

AI-Integrated healthcare as a solution may present us the potential breakthrough for the aforementioned challenges. Emerging research suggests that large language models (LLMs), trained via autoregression, supervised fine-tuning, and human preference optimization, now approach human-level performance in multifaceted tasks (e.g., content summarization [3,4], medical Visual QA [5], and mathematical reasoning[6,7]). Yet, in high-stakes medical domains, even specialized models (e.g., GeneGPT [8], BioGPT [9], AlphaFold[10], Tianyi, TCM-Chat) exhibit critical gaps in diagnostic accuracy and clinical decision reliability. Besides, Current medical AI approaches—including task-specific LLMs (e.g., ICD coding prediction (classification) [11], herbal prescription recommendation or generation (classification/generation) [12-16])—fail to capture the holistic nature of elite physicians' specialized knowledge systems[17]. A master physician's expertise is not a mere aggregation of classification/generation tasks, but a comprehensive framework which originates from the physician's mastery and comprehension of foundational medical knowledge, evolves through their reflective refinement of diagnostic principles in response to the medical knowledge system, and matures via continuous reconciliation of theoretical knowledge with clinical practice. Ultimately, it crystallizes into a personalized methodology for diagnosing and treating a broad category of diseases. To authentically replicate this, AI systems must interconnect diverse medical knowledge through reasoning-aligned architectures, constructing internal logic pathways that mirror the target physician's decision-making. Only such anthropomorphic modeling could emulate expert-level practice in real-world scenarios.

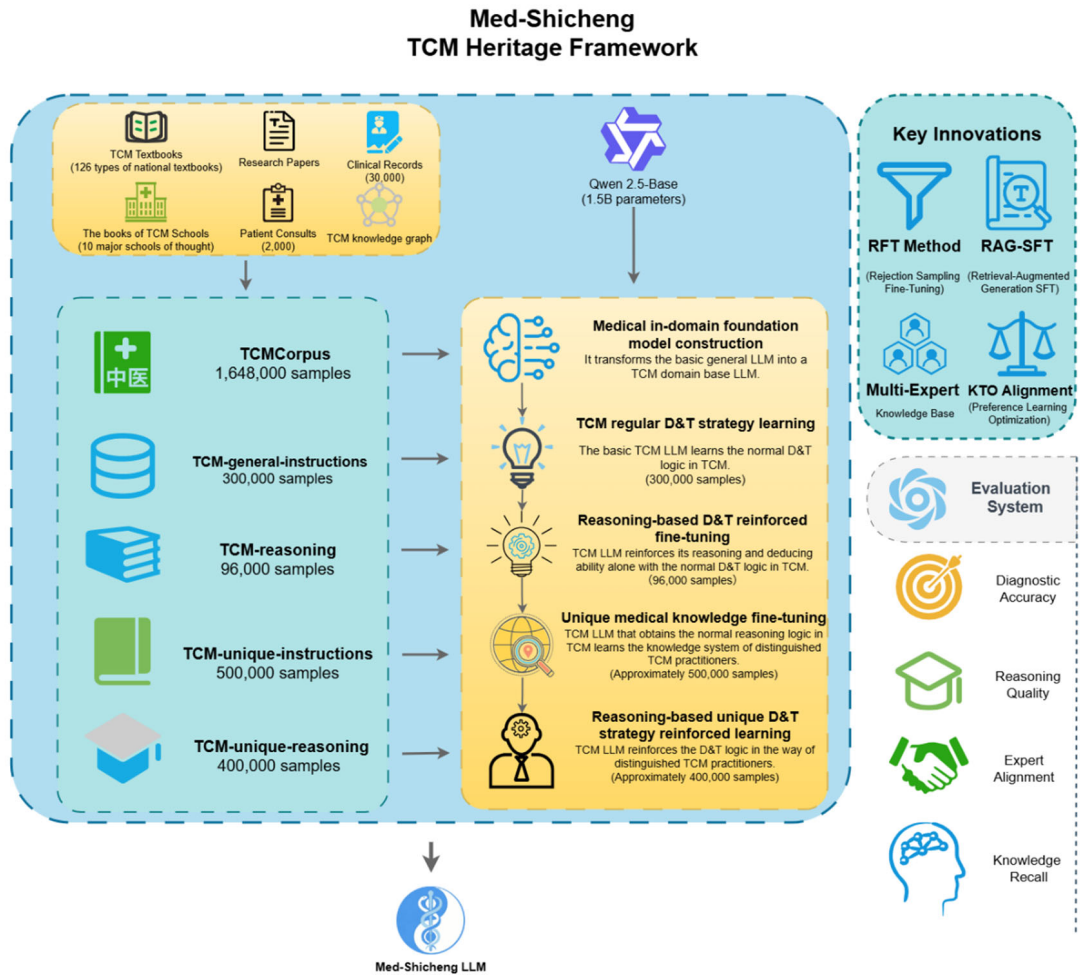

Fig. 1 In this work, we proposed a general paradigm to enable the LLM learning the outstanding medical knowledge system and D&T experience of the distinctive TCM doctors, named Med-Shicheng. It consists of five stages to ensure a general-purpose LLM into a medical domain LLM that can master the D&T knowledge and strategies of different distinctive TCM doctors: (1) Medical in-domain foundation model construction, (2) TCM regular D&T strategy learning, (3) Reasoning-based D&T reinforced fine-tuning, (4) Unique medical knowledge fine-tuning, and (5) Reasoning-based unique D&T strategy reinforced learning. In each stage, we design and construct a dataset according to the TCM domain tasks and logic. To evaluate the model's performance trained by Med-Shicheng, we also design an evaluation framework to evaluate all comparison models from four dimensions: diagnostic accuracy, reasoning quality, extent of expert alignment, and knowledge recall.

In this study, we propose a general framework, named **Med-Shicheng[b]**, that enable LLM to learn various type of knowledge mentioned above systematically. Then, we applied it to learn the experience and knowledge of 5 National Master of Chinese Medicines or distinguished TCM physicians as a representative example, showing the effectiveness and efficiency of Med-Shicheng. The framework is shown in Fig.1. In this application, the model is requested to generate for 7 TCM tasks with the target TCM distinguished physicians' knowledge systems simultaneously, including **TCM knowledge–based analysis of etiology and pathogenesis**, **syndrome diagnosis**, **analysis and determination of treatment principles**, **TCM prescription generation**, **prescription**

---

[b] In Chinese cultural discourse, Shicheng (师承) represents a paradigm of knowledge transmission wherein cultural practices, artisanal techniques, and philosophical wisdom are systematically preserved and adapted across generations.

**explanation**, **potential symptom changes after medication and herbal modifications**, and **medical advice**. Med-Shicheng is built on the basis of Tianyi [18], and comprises five stages: (1) medical in-domain foundation model construction, (2) TCM regular D&T strategy learning, (3) reasoning-based D&T reinforced fine-tuning, (4) unique medical knowledge fine-tuning, and (5) reasoning-based unique D&T strategy reinforced learning. To complete transmission of target knowledge system, various types of resources that presents the clinical philosophy and therapeutic paradigm, medical intellectual tradition, and clinical D&T reasoning logic are collected, and be exploited in the aforementioned stages. The research articles of distinguished TCM physicians usually summarized and verified by their students. They include the systematical D&T knowledge over a kind of disease or a group of representative symptoms, and how such knowledge applied in clinical cases, which delivers the most authoritative and comprehensive knowledge system of target physicians. The clinical case portfolios of target physicians constitute primary-source, real-world evidence documenting the application of their characteristic clinical reasoning logic for syndrome differentiation and personalized therapeutic intervention. These two types of contents are sufficiently comprehensive and methodical to encapsulate the D&T philosophy and practices of master TCM practitioners, with the primary challenge lying in their optimal utilization. To fully leverage these resources, we design different tasks in the five stages and organize these resources in various format accordingly. The samples of all distinguished TCM practitioners for each stage are utilized simultaneously. Besides, we implement Med-Shicheng with Qwen2.5-1.5B-Base [19] in this study, while keeping its performance is comparable with the biggest version of Deepseek-R1 [20].

The contributions we made in this study are following. **First**, we propose a general framework, named **Med-Shicheng<sup>c</sup>**, for the standardized transmission of personalized medical knowledge system of distinguished doctors in a unified way. It enables the highly efficient integration of both labeled and unlabeled data, effectively harnessing their information in a mutual manner. **It achieves performance that is competitive with general-purpose LLMs by a lightweight LLM**, despite being trained on a very small number of labeled samples—**merely hundreds of valid original cases in this study. Second**, in this work, the **Qwen2.5-1.5B-Base model is utilized, implemented within the Med-Shicheng framework, and delivers performance on par with DeepSeek-R1-671B despite its 447x smaller parameter count**, which can be trained and deployed in on resource-constrained GPUs, such as the consumer-grade GPUs. It proves that the LLM with very limited model scale can achieve the outstanding performance and solve the complicated tasks in specialized domain if researchers can design the tasks according to the in-domain issues, showing a promising potential in in-domain LLM researches in a cost-effective manner. **Third**, we propose an effective strategy to integrate diverse medical knowledge systems into a single model, avoiding misleading interference or negative interactions between them. This approach eliminates the need for labor-intensive efforts to transfer the knowledge of distinguished medical physicians into separate individual models. **Fourth**, we provide a detailed analysis on the feasibility of LLMs-as-Judges paradigm for the evaluation of informative long responses generated by the professional in-domain models. Based on our analysis, we propose that involving human physicians in model assessment is indispensable when corresponding ground truth is unavailable even though current state-of-the-art LLMs, such as GPT-5, Deepseek-R1, and Gemini, are capable of evaluating such responses to some extent.

---

<sup>c</sup> The introduction of Med-shicheng can be found at https://njucm-bjucm-tcm-ai.github.io/Med-Shicheng_project_website/

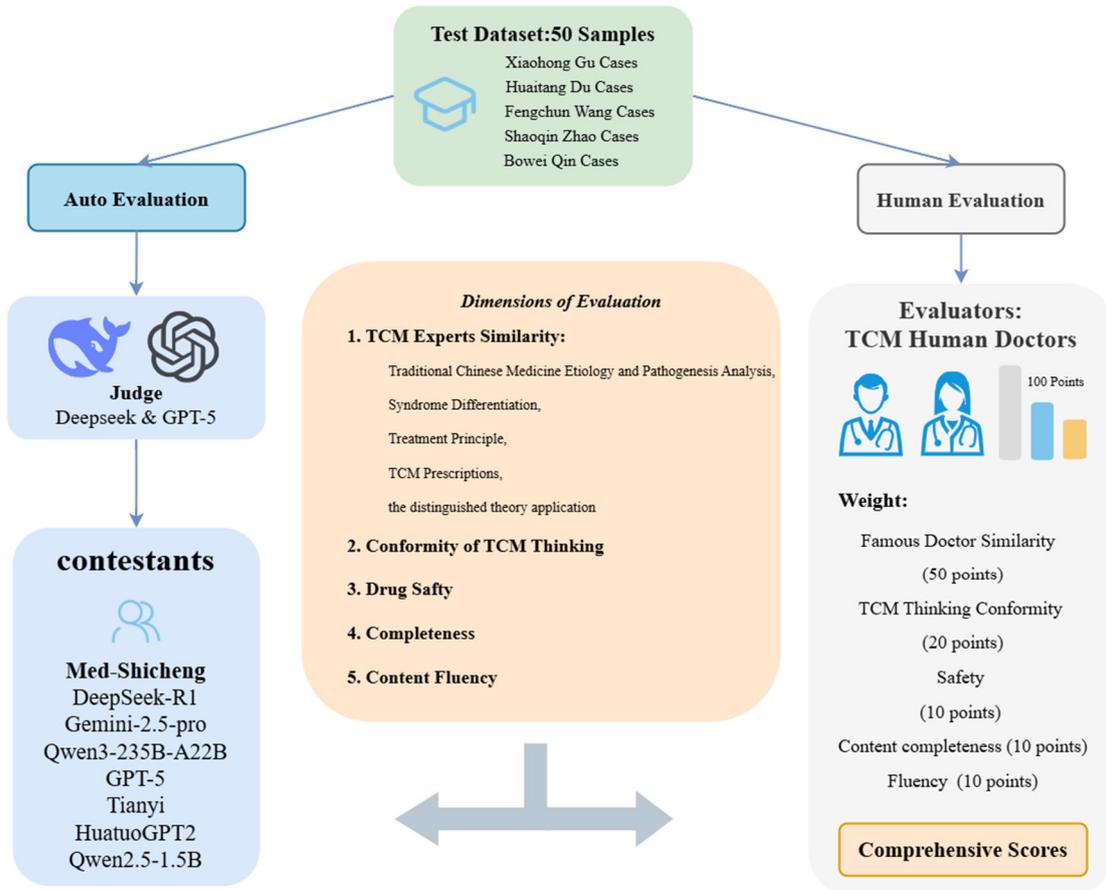

Fig 2. The evaluation Framework. It contains two types of evaluations: automatic evaluation and Human doctor evaluations. For the automatic evaluation, we take the state-of-the-art LLMs with universal complex problem-solving abilities such as Deepseek and GPT as the judge to evaluate all comparison models' prediction on test samples. In this study, we take both Deepseek-V3.2 and GPT-5 as the judges which means each model generated response will be evaluated twice. Deepseek-V3.2 and GPT-5 assess if the responses generated from all comparison models are desirable for the query by their own and considering the golden label at the meantime. The reason we choose this evaluation method is that the response to the tasks in this study is a long text over the test clinical case which contains seven essential parts with professional description to express the same meaning, so does the label. It's impossible to evaluate them by the classic evaluation metrics, such as accuracy, recall, and F1-score. The higher score model obtained, the better performance of the corresponding model. For the human doctor evaluation, we invited 18 experienced TCM doctors with senior title, and divide them into five groups of TCM doctors for these five targeted distinguished TCM doctor, all of whom are the apprentices or students in each group. We employed the Delphi method to establish TCM expert consensus standards, offering the invited TCM doctors a unified evaluation principles and requirements to evaluate all comparison models from five dimensions: (1) the D&T similarity to target distinguished TCM doctors, (2) the consistency with TCM philosophical framework, (3) the safety compliance, (4) therapeutic completeness, and (5) clinical coherence. The detailed information about human expert evaluation standards can be found in **Appendix B**.

## Evaluations

We design two types of evaluations to assess the performances of all comparison models, i.e., the automation evaluation and the TCM human doctor evaluation, of which the first assess with the quantitative evaluation without human bias, and the second evaluates the responses of all models by the professional human experts in TCM logical thinking. The whole evaluation framework is shown in Fig. 2.

## Comparison Baselines

Considering the complexity of the tasks in this study, we choose the LLMs that has the complicated-problem-solving ability. The **Deepseek-R1**, **GPT-5**, **Gimini-2.5-pro**, and **Qwen3-235B-A22B-Thinking** are proved to have the SOTA reasoning and solving performance on various complicated tasks. The models of **HuatuoGPT2-7B** series are evaluated in our previous work and show its solid performance on TCM tasks among various TCM LLMs. **Qwen2.5-1.5-instruct** is also selected due to the same series of the base model of Med-shicheng. Thus, we choose to use Deepseek-R1, GPT-5[21], Gimini-2.5-pro [22], Qwen3-235B-A22B-Thinking [23], Qwen2.5-1.5-instruct [19], and HuatuoGPT2-7B [24] as the comparison baselines in this study.

## Automation Evaluation Strategy

To evaluate the responses generated by all comparison LLMs according to the test TCM clinical cases without human biases, we design the automatic evaluation. Different from the normal classification or generation tasks, the responses generated by all comparison LLMs are the token sequences which are more than 1500 tokens. Each sequence contains *the TCM etiology and pathogenesis analysis*, *syndrome differentiation*, *treatment principles*, *TCM prescription, prescription explanation*, *application of academic thought of the distinguished doctor*, and *symptom changes and medication adjustments*. Each part among the seven tasks contains at least a paragraph to describe the corresponding content, which is hard to evaluate its correctness with the classic evaluation metrics such as accuracy, precision, recall, and F1-score. The metrics of Bleu and Rouge are also proved unable to evaluate the response generated by LLM [25,26]. Thus, to overcome these challenges, we design a detailed and comprehensive prompt that split each response from a comparison LLM and the golden label along with the seven parts mentioned above. Then, we take the state-of-the-art general-purpose LLM, which are the Deepseek-V3.2 and GPT-5 in this study, as the judges to evaluate each item's correctness by their own and considering the golden label simultaneously. Finally, all seven scores are normalized and compute the average score. The higher score represents the better model response for the TCM clinical case. The prompt we designed for Deepseek-V3.2 and GPT-5 can be found in the section 1 *Appendix B*.

## Human Evaluation Strategy

Although the automation evaluation has no human subjective biases, its evaluation quality relies on the underling model that extract the representation for each item. Automation evaluation can only deliver coarse-grained insights at the holistic semantic level, which may lead to less rigorous assessments compared to human expert analysis. **Thus, we employed the Delphi method[27] to establish TCM expert consensus evaluation standards for LLMs**. In human evaluation, we invited 18 TCM experience doctors with senior titles and divide them into five groups. These TCM doctors are either the apprentices or students of the target distinguished TCM doctor, and required to evaluate responses of the TCM clinical cases from corresponding distinguished TCM doctor according to the TCM expert consensus evaluation standards for LLMs we proposed. The items in the evaluation standards contains *the similarity to distinguished physicians* (50 points), *the consistency with TCM philosophical* (20 points), *safety* (10 points), *content completeness* (10 points), *fluency* (10 points). The rating form TCM doctors used can be found section 2 of *Appendix B*

# Experiments

## Automatic Evaluations

We try to design an objective strategy with the well-defined and fine-grained prompt to assess all comparison models' performances without human bias. After trying several general-purpose state-of-the-art (SOTA) LLM, we select the Deepseek-V3.2 and GPT-5 with the most comprehensive evaluation capacities as judges. The results of Deepseek-V3.2 and GPT-5 are illustrated in the part a. and part b. of Fig. 3.

**Evaluations by Deepseek-V3.2.** As we can noticed in part a, all comparison models express a consistent performance trend. An exception occurs for the scores obtained by HuatuoGPT2-7B on doctor Huaitang Du's test clinical cases. The *TCM prescription* and *treatment principle* scores of HuatuoGPT2-7B are the highest among all comparison models while the rest items' scores are extremely poor. Except for above, HuatuoGPT2-7B obtains the two lowest-ranking scores among all the rest target doctor's cases. To verify this anomaly phenomenon, we review the prediction of HuatuoGPT2-7B on all Dr. Huaitang Du's cases. We noticed that it generated the identical TCM prescriptions and treatment principles with the labels over half of the cases while the contents of the rest items are rarely correct. Considering all performances of HuatuoGPT2-7B on all doctors' cases, these exceptional scores illustrate that some Dr. Huaitang Du's test cases might be included in the training data of HuatuoGPT2-7B.

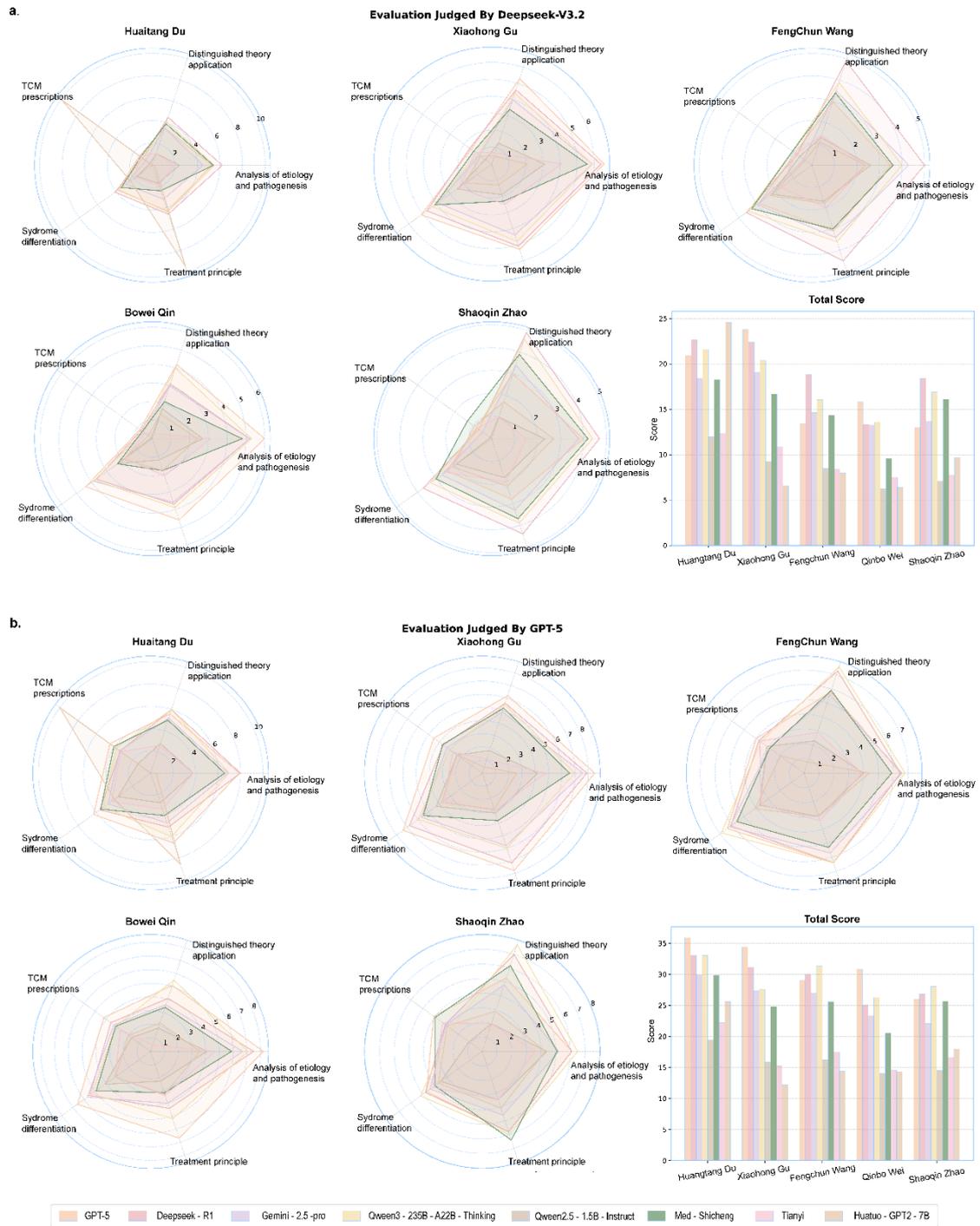

Fig. 3 The automatic evaluation results judged by Deepseek-V3.2 and GPT-5, which are illustrated in subfigure a. and b. respectively. In general, the evaluation results by Deepseek-V3.2 and GPT-5 shows the consistent trend for all model's prediction on all test samples of target distinguished TCM doctors except for the scores for HuatuoGPT2-7B on doctor Huaitang Du's test samples which obtains the highest TCM prescription and treatment principle results. By reviewing HuatuoGPT-7B's genearion on Dr. Huaitang Du's test samples, we noticed that it achieved exceptionally high scores in TCM prescription and treatment principles by generating content identical to the labels in half of the cases, while differing significantly in other items. This anomaly suggests that these specific samples may have been part of HuatuoGPT2-7B's training data. For total scores, the comparison models can be divided into two distinct groups: (1) the higher performance models: Med-Shicheng, GPT-5, Deepseek-R1, Qwen3-235B-A22B-

Thinking, Gemini-2.5-pro; and (2) the lower performance models: Qwen2.5-1.5B-instruct, Tianyi, HuatuoGPT2-7B. **The Med-Shicheng is the only model, which has the least parameters, that can generate the comparable quality content to the models with hundreds of billions of parameters.** However, there is no model that can obtain the consistently highest scores on all target doctor's test samples. Taking the results of Deepseek-V3.2 as an example, with the largest model scales, Deepseek-R1 and GPT-5 obtain the top 2 scores in most items of most cases, while Med-shicheng, Gemini-2.5-pro, and Qwen3-235B-A22B-Thinking also show their desirable performance than Deepseek-R1 and GPT-5, such as in the case of Fengchun Wang, Shaoqin Zhao, and Huaitang Du. The scores judged by GPT-5 also confirm these trends with the larger evaluation values than Deepseek. All scores are the average scores across all test samples.

Generally speaking, Med-shicheng, GPT-5, Deepseek-R1, and Qwen3-235B-A22B-Thinking obtains the top-4 performances among all comparison models. However, there is no model that can achieve the consistent highest scores among all target doctor's cases. Deepseek-R1 obtains the highest scores in both Huaitang Du, Fengchun Wang, and Shaoqin Zhao on *distinguished theory application*, *analysis of etiology and pathogenesis*, *treatment principle*, and *syndrome differentiation* items. Qwen obtains the highest scores on these items at Dr. Xiaohong Gu' cases, while GPT-5 achieves the similar status on Dr. Bowei Qin's cases. The Med-shicheng obtains the comparable scores with these large-scale LLMs, such as, the Dr. Huaitang Du, Xiaohong Gu, Fengchun Wang, and, Shaoqin, Zhao. It's also worthy to notice that the performances of Med-shicheng are identical even better than Gemini-2.5-pro and GPT-5, such as the scores of Med-shicheng on all evaluated items are better than Gemini-2.5-pro's on Dr. Shaoqin Zhao's cases; and the same phenomenon occurs between Med-shicheng and GPT-5 on Dr. Fengchun Wang with only exception at treatment principle which are 3 for Med-shicheng and 3.1 for GPT-5. The Med-shichang also obtains consistently comparable scores on all items, even the best score on TCM prescription item. For instance, Med-shicheng demonstrates the consistent high scores on *distinguished theory application*, *analysis of etiology and pathogenesis*, and *syndrome differentiation* by both Deepseek-V3.2 and GPT-5, which shows the stable and desirable reasoning ability of Med-shicheng with only 1.5 billion parameters over the reason why the current symptoms occur through the target TCM knowledge system. According to the training strategy and the datasets, this ability is mainly built on the sufficient TCM knowledge pre-training and the well-designed strategy to learn the target distinguished TCM doctor's special knowledge system, and also benefits from that the special knowledge systems stem from the same theoretical root.

For the Qwen2.5-1.5B-instruct, Tianyi, and HuatuoGPT2-7B, both of these three LLMs obtain the total scores that less than 10, which shows their limited abilities on mastering distinguished TCM doctor's special knowledge system no mater of the small or large parameter scale (from 1.5B to 7B). Among these three models, we can observe that Tianyi model obtains the best total scores in most target doctors' cases except for the abnormal scores of HuatuoGPT2-7B. As the two TCM specialized LLM, Tianyi achieves the highest scores on all items at cases of Dr. Xiaohong Gu, Fengchun Wang, Bowei Qin. The only exception occurs on *distinguished theory application* item at Dr. Bowei qin. A promising performance is achieved by Qwen2.5-1.5B-instruct. It achieves the better total scores and item scores in most cases than HuatuoGPT2-7B at Dr. Xiaohong Gu, and Fengchun Wang, and the comparable total and item scores at Dr. Bowei Qin, and Shaoqin Zhao, which shows the its ability to master the special TCM knowledge while such ability is not consistency in all cases. However, such ability with general-purpose LLM is somehow better than the specialized trained TCM LLM, indicating the effectiveness of sufficient and efficient training strategies and methods are helpful and hopeful for domain-specialized LLM construction.

**Evaluations by GPT-5.** For the results from GPT-5, we observe the similar characteristics and trends among target distinguished TCM doctors with the results from Deepseek-V3.2. For instance, all comparisons LLMs are also divided into two groups: (1) the Med-shicheng and 4 general-purpose LLMs of which each boasting in excess of 100 billion parameters; (2) Qwen2.5-1.5B-instruct, Tianyi, and HuatuoGPT2-7B. The Med-shicheng achieves the lowest performance on Dr. Bowei Qin's cases.

Table 1. The scores differences between the Med-shicheng and GPT-5, Deepseek-R1, Gemini-2.5-pro, and Qwen3-235B-A22B-Thinking. It displays the score differences (Δ) between each model and the Med-shicheng benchmark, calculated as (LLM) – (Med-shicheng). Negative values are shown in parentheses. The analysis focuses on identifying bias, especially in GPT-5 and Deepseek-V3.2.

| *Huaitang Du* | GPT-5 scores Δ | | | | Deepseek-V3.2 scores Δ | | | |
|---|---|---|---|---|---|---|---|---|
| model | GPT-5 | Deepseek-R1 | Gemini-2.5-pro | Qwen3-235B-A22B-Thinking | GPT-5 | Deepseek-R1 | Gemini-2.5-pro | Qwen3-235B-A22B-Thinking |
| Δ Total | 2.756 | 2.863 | (0.59125) | 2.40675 | 5.299 | 4.576 | 3.524 | 3.745 |
| Δ Analysis of etiology and pathogenesis | 1 | 0.8 | (0.3) | 0.9 | 1.4 | 1.2 | 0.6 | 0.5 |
| Δ Syndrome differentiation | 0.6 | 0.6 | (0.3) | 0 | 1 | 0.7 | 0.3 | 0.5 |
| Δ Treatment principle | 1.9 | 1.5 | 1 | 1.4 | 1.8 | 1.5 | 1.2 | 1.3 |
| Δ TCM Prescriptions | 0.056 | (0.237) | (0.59125) | 0.10675 | 0.099 | (0.124) | (0.076) | (0.055) |
| Δ Distinguished theory application | (1.1) | 0.8 | (0.7) | (0.3) | 0.8 | 0.8 | 1 | 0.5 |

The most significant difference between GPT-5 and Deepseek-V3.2 lies in their scoring styles. GPT-5 systematically assigns higher scores across all evaluation items, whereas Deepseek-V3.2 demonstrates a consistently lower scores tendency. Such scoring behaviors compensated for the outlier issue associated with HuatuoGPT2-7B. Since GPT-5 generally assigns higher scores across all items, HuatuoGPT2-7B's notably poor performance in areas other than *TCM prescriptions* and *treatment Principles* prevented its total score from being disproportionately high. Consequently, the overall score comparison clearly revealed the HuatuoGPT2-7B's performance deficiencies—a distinction that the Deepseek-V3.2 scoring results failed to capture. For instance, we can observe that HuatuoGPT2-7B's total score does not surpass the LLMs in group (1). The other difference includes: GPT-5 tends to assign all LLMs with higher scores on TCM prescription item while Deepseek-V3.2 is not. The GPT-5 assigns itself generations the highest scores on all doctors' cases except for Fengchun Wang and Shaoqin Zhao, while Deepseek-V3.2 does not show such preference.

To investigate potential scoring bias and its extent—particularly regarding GPT-5 and Deepseek-V3.2—we calculated the score differences between the benchmark model (Med-shicheng) and the other models. The table reports these differences, symbolized by Δ, which are computed as (Model Score) - (Med-shicheng Score). A negative difference is indicated by parentheses. We only present the scores differences on Dr. Huaitang Du's cases due to the length limitation. All other doctors' score difference can be found in ***Appendix C.*** As shown in Table 1, regardless of whether we consider Deepseek-V3.2 or GPT-5, the score differences are usually large than 1 at *analysis of etiology and pathogenesis*, *treatment principle*, *distinguished theory application*, and *total* items, while the differences on *syndrome differentiation* and *TCM prescriptions* are very small. The score differences on other doctors' cases demonstrate the similar trends. Thus, with different model architectures, training strategies, and datasets of Deepseek-V3.2 and GPT-5, both SOTA LLMs illustrate the similar trends of score differences, indicating the less evaluation bias of them on Med-shicheng's generations.

We further investigated factors influencing Med-shicheng's performance by jointly analyzing data scale, data type, and the results illustrated in Fig. 3. The performance on each distinguished TCM doctor's clinical cases suggest that it is largely influenced by the scale of the target doctor's own clinical cases and specialized TCM knowledge, under a consistent training framework. The numbers of valid real-world clinical cases for the target doctors are as follows: Huaitang Du (872), Xiaohong Gu (570), Fengchun Wang (140), Bowei Qin (280), and Shaoqin Zhao (160). The tokens constructed from each doctor's specialized TCM knowledge system (books and articles) are: Huaitang Du (64,503), Xiaohong Gu (58,333), Fengchun Wang (188,833), Bowei Qin (60,102), and Shaoqin Zhao (185,252). Med-shicheng achieves comparable or even optimal performance when either the valid clinical cases or the specialized TCM knowledge tokens are sufficiently large for a specific doctor. For example, as shown in Fig. 3, the model performs comparably to SOTA LLMs on cases from Dr. Huaitang Du, Xiaohong Gu, Fengchun Wang, and Shaoqin Zhao—doctors with either ample clinical cases (over 500 for Du and Gu) or extensive knowledge tokens (around 180,000 for Wang and Zhao). In contrast, performance is relatively lower for Dr. Bowei Qin, where both clinical cases and knowledge tokens are limited compared to the others. Thus, it can be concluded that, under our proposed training framework, valid clinical cases and specialized medical knowledge mutually enhance model performance, even with limited target-labeled samples. This also demonstrates that the Med-shicheng framework effectively reduces the dependency on costly labeled data construction—which typically requires hundreds of thousands of reasoning and non-reasoning samples—thus lowering the associated expenses. Besides, the valid clinical cases and specialized medical knowledge token demonstrates different efficiency on improving model's performance—increasing the number of clinical cases contributes more substantially to model's performance than expanding the volume of specialized medical knowledge tokens.

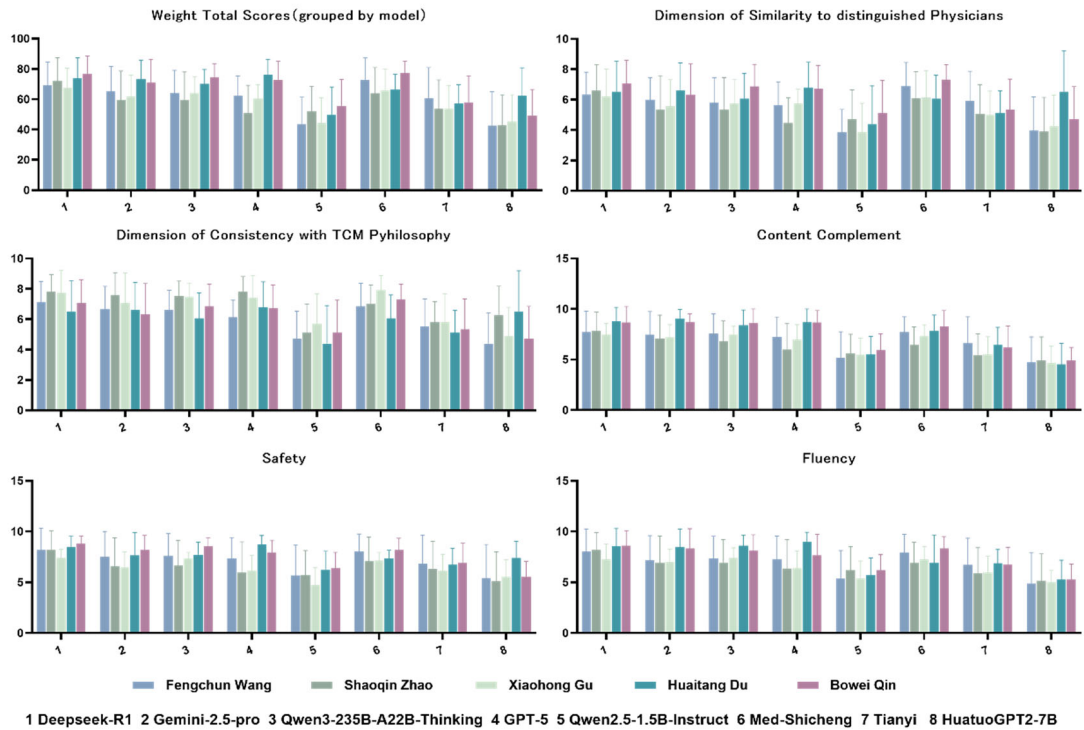

Fig. 4. The human doctor evaluations for all target distinguished TCM doctors. All scores of comparison models are grouped by doctors. The weighted total scores that are grouped by models are displayed separately in a sub-figure. For the clearness, we named all comparison models' name with the numbers from 1 to 8, and their names are placed at the bottom of the figure. All target distinguished doctors' cases evaluations are grouped by model, and each of the evaluation dimension are presented in a sub-figure. Overall, human doctors tend to offer the different rating distributions compared to GPT-5 and Deepseek-V3.2. Human doctors prefer score Med-shicheng with higher scores for most target TCM doctors' cases in almost all 5 dimensions while for results presented in Fig.3 GPT-5 and Deepseek-V3.2 are not. For instance, Med-shicheng obtains the scores comparable to the top or even the top scores in the cases of HuaiTang Du, Xiaohong Gu, Fengchun Wang, and Bowei Qin on dimension of similarity to distinguished physicians with the small variances. However, in Fig. 3., the GPT-5 and Deepseek-V3.2 merely rate Med-shicheng with top scores. This pheromone may propose that the LLMs with extremely large parameters scale (e.g., GPT-5 or Deepseek-R1) are capable of evaluating the complicated and long content generated by specialized trained in-domain LLMs under the elaborately designed instructions. Yet, such ability is limited when the evaluation requirements are becoming more specific and professional. In this work, the LLMs are required to assess if the comparison models can generate the response along with the target doctor's habits in TCM diagnosis, treatment principles, medication. They only assess these contents with the ordinary TCM knowledge and provide the scores even though for the cases of Huaitang Du whose books and resources are widely accessible. While, human TCM doctors can differentiate the differences of doctor's specialized TCM system and the normal TCM knowledge presented in models' responses. Such evaluation weakness of GPT and Deepseek is more obvious in the scores of Bowei Qin, human doctors rated Med-shicheng at the second place in almost 5 dimensions in cases of Bowei Qin while GPT and Deepseek rated it at the fourth place. The similar evaluation differences occur in the cases of HuaiTang Du and Xiaohong Gu.

## Human Evaluations

In this evaluation, we invited 18 experienced TCM doctors to assess the responses generated by all comparison models. All TCM doctors are given the model responses and its rating forms, and required to rate each model's responses in five dimensions, i.e., **Dimension of Similarity to distinguished Physicians**, **Dimension of Consistency with TCM Philosophy**, **Safety**, **Content Completeness**, **Fluency**. **All models' name are blind**, and more detailed information about the evaluation framework can be found in the table 2 of **Appendix B**. Due to the fact that three distinguished physicians (Wang Fengchun, Zhao Shaoqin, and Qin Bowei) are deceased, and the other two (Gu Xiaohong and Du Huaitang) are advanced in age (Professor Du Huaitang is over 90 years old, and Professor Gu Xiaohong is over 60), conducting a complete manual evaluation of all medical cases would require nearly a week of intensive, focused assessment. Therefore, we engaged their successors and disciples to perform the quantitative evaluation. The qualitative assessment of the case records was performed by the target physicians themselves, where still living. For instance, regarding the evaluation of cases from Professor Gu Xiaohong, while she did not assess all cases, she conducted a blinded qualitative evaluation of the content generated by eight models for randomly selected cases. Since Professor Gu Xiaohong did not complete the scoring for all cases, her evaluation results are used solely for qualitative analysis. The titles and number of evaluators involved in the quantitative assessment are as follows: **Zhao, Shaoqin Cases:** 4 evaluators in total (2 with Senior Title, 1 with Associate Senior Title, 1 with Intermediate Title). **Wang, Fengchun Cases:** 4 evaluators in total (2 with Senior Title, 1 with Associate Senior Title, 1 with Intermediate Title). **Qin, Bowei Cases:** 3 evaluators in total (1 with Senior Title, 1 with Associate Senior Title, 1 with Intermediate Title). **Du, Huaitang Cases:** 3 evaluators in total (1 with Senior Title, 1 with Associate Senior Title, 1 with Intermediate Title). Gu, Xiaohong Cases: 3 evaluators in total (2 with Associate Senior Title, 1 with Intermediate Title). The human evaluations over all target TCM doctors are shown in Fig. 4.

Overall, the rating distributions provided by human doctors differ markedly from those produced by GPT-5 and Deepseek-V3.2. Human experts tend to assign Med-Shicheng higher scores than the other models for most target TCM physicians and across almost all five evaluation dimensions, whereas GPT-5 and Deepseek-V3.2 do not show this preference (see Fig. 3). For example, on the dimension of *similarity to the distinguished physician*, Med-Shicheng receives the nearly highest scores in the cases of Xiaohong Gu ($6.13\pm1.77$ while $6.2\pm1.82$ for Deepseek-R1), ShaoQin Zhao ($6.1\pm1.734$ while $6.6\pm1.7$ for Deepseek-R1) or even highest scores in the cases of Fengchun Wang ($6.9\pm1.55$), and Bowei Qin ($7.33\pm0.98$), with only small variances among human raters. In contrast, GPT-5 and Deepseek-V3.2 rarely rank Med-Shicheng at the top in Fig. 3. This pattern suggests that very large LLMs (e.g., GPT-5 and Deepseek-V3.2) can evaluate long and complex outputs from in-domain models under carefully designed instructions (see section 1 of Appendix B), but their capability is limited when the evaluation criteria become highly specific and professional. In this study, the LLMs are asked to judge whether each comparison model's responses follow the target physician's habits in TCM diagnosis, therapeutic principles, and medication use. GPT-5 and Deepseek-V3.2 appear to rely mainly on general TCM knowledge when assigning scores, even for cases such as Huaitang Du, whose books and materials are widely available. Human TCM physicians, by contrast, can distinguish between responses that merely reflect standard TCM theory and those that embody the unique theoretical system and prescribing style of a particular master. Such phenomenon also appears in the scores of HuatuoGPT2-7B. We can also notice that HuatuoGPT2-7B obtains the exact same score with Deepseek-R1 at this dimension with very little difference on variance. It confirms the results provided by automation evaluations in Fig. 3., which is that HuatuoGPT2-7B trained on the resources that contains lots of cases and text of Huaitang Du. Yet, its scores on the other dimensions are not as significant as the dimension of *similarity to the distinguished physician*, which demonstrates the limited model performance and lower training efficiency on the TCM resources on the modeling of the specialized TCM knowledge system. Such evaluation weakness of GPT-5 and Deepseek-V3.2 is especially evident in the cases of Bowei Qin: human doctors place Med-Shicheng second on almost all five dimensions, whereas GPT-5 and Deepseek-V3.2 rank it fourth. Similar discrepancies are observed in the cases of Huaitang Du and Xiaohong Gu. Unexpectedly, on the two most critical dimensions *similarity to distinguished physicians* and *consistency with TCM philosophy*, Med-shicheng achieves the highest scores with minimal variance in the cases of Bowei Qin and Fengchun Wang, while securing second place in the cases of Shaoqin Zhao and Xiaohong Gu. These evaluation results differ significantly from those of GPT-5 and Deepseek-V3.2, highlighting the limitations of using extremely large LLMs for such specialized assessment tasks.

Except the abnormal scores of HuatuoGPT2-7B on cases of Huaitang Du, the score distributions of the LLMs that have not very large parameters provided by human TCM doctors are similar to the score distributions of GPT-5 or Deepseek-V3.2. For instance, Tianyi achieves the best scores among Qwen2.5-1.5B-instruct and HuatuoGPT2-7B on all 5 evaluation dimensions with relatively small variances. Qwen2.5-1.5B-instruct and HuatuoGPT2-7B have their advances on different dimensions of different target distinguished TCM doctors' cases, but formed the consistent advantages on in all. For example, as a general-purpose LLM, Qwen2.5-1.5B-instruct achieves the competitive even better scores than HuatuoGPT2-7B at the dimension of *similarity to the distinguished physician* on the cases of Fengchun Wang, Shaoqin Zhao, and Bowei Qin, while HuatuoGPT2-7B obtains the better performance on the cases of Xiaohong Gu. Compared with Med-shicheng and very Large LLMs, these 2 models usually obtain the relatively large variances on these cases.

An interesting result appears on Tianyi. Without any supervised fine-tuning and reinforced learning, Tianyi receives the comparable performance on the cases of Fengchun Wang, Shaoqin Zhao, and Xiaohong Gu with GPT-5 and Qwen3. For example, for cases of Shaoqin Zhao, Tianyi receives $5.05\pm1.93$, $5.35\pm2.08$, $5.45\pm2.11$, $6.3\pm2.74$, $5.9\pm2.51$ for dimensions of *similarity to distinguished physicians, dimension of consistency with TCM philosophy, safety, content completeness, fluency*, respectively, which is better than the GPT-5's performances on *similarity to distinguished physicians* ($4.45\pm1.67$), *dimension of consistency with TCM philosophy* ($5.1\pm2.43$), *safety* ($6\pm2.99$). The similar results appear on Fengchun Wang and Xiaohong Gu that is the scores Tianyi receives are either better than the GPT-5 or Qwen3 on *similarity to distinguished physicians* or very close to them on the other dimensions. Med-shicheng and Tianyi are two models with high relevance in our study. It demonstrates the advanced efficiency of our proposed strategy on modeling in-domain expertise and knowledge system. After human TCM doctors finished their evaluations over all models, we conduct a brief interview for each of them. A consistent agreement held by all human TCM doctors is that while some general-purpose LLMs (e.g., DeepSeek, Gemini) demonstrate rigorous analytical logic and align with TCM reasoning in their expressions, their content still shows significant discrepancies with the physicians' specialized diagnostic philosophies and medication habits. Additionally, these models tend to generate redundant and verbose content. In contrast, the Med-Shicheng demonstrates high comfortability in its writing style, content length, diagnostic philosophy, and medication habits, aligning more closely with the conventional expression standards of TCM case records.

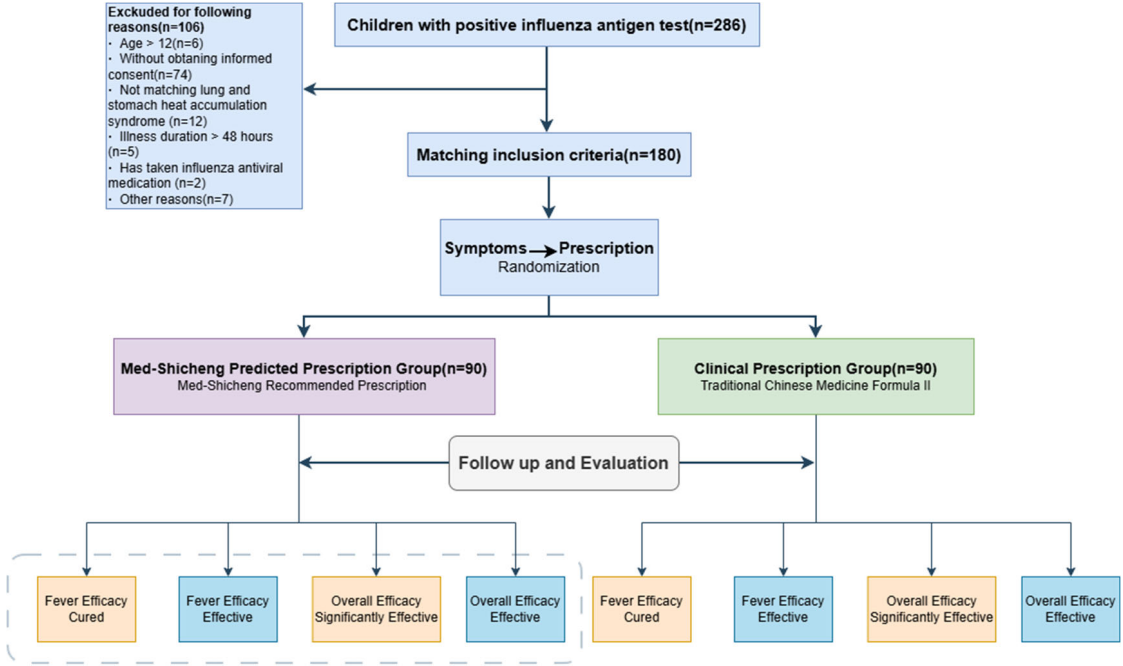

Fig. 5. Patient enrollment strategy and model predictions for TCM clinical trials.

**Traditional Chinese Medicine Clinical Trial Prediction**.

In this task, each LLM predicts the result for the target clinical trial task according to its task definition and clinical case information. The methods for enrolling patients in this TCM clinical trials can be found in Fig. 5. Prior to its initiation, this study was registered at the Chinese Clinical Trial Registry (ID: ChiCTR2100054426, date: 2021.12.17) and received approval from the Medical Ethics Committee of Beijing University of Chinese Medicine (ID: 2021BZYLL1024)[28]. This is a clinical trial to verify the effectiveness of the TCM prescription from Dr. Xiaohong Gu. In this retrospective study, our goal is to verify whether Med-shicheng can generate the similar TCM prescription with the target trial TCM prescription since it had been learned the Dr. Xiaohong Gu' cases and specialized knowledge system. Thus, the models compared in this retrospective study of clinical trial are required to generate a TCM prescription according to the patients' medical history, current symptoms, and physical signs. This retrospective study is designed by the zero-shot learning strategy, all valid patients' records are used to evaluate the compared models. We calculate the overlapped herbs between model-generated TCM prescription and the label as the evaluation metric to indicate the extent of the model master Dr. Xiaohong Gu's specialized knowledge, the higher similarity the better performance of a model on this study. All test samples' average as the final score, and we also provide the standard variance to assist further analysis.

Table 2. The average scores and its variances of all comparison models that can generate the valid response with the instructions and the query of each sample in clinical trial.

| model | GPT-5 | Deepseek-R1 | Gemini-2.5-pro | Qwen3-235B-A22B-Thinking | Tianyi | Qwen2.5-1.5B-instruct | **Med-Shicheng** |
|---|---|---|---|---|---|---|---|
| score | 38.41 | 36.77 | 44.50 | 39.31 | 17.15 | 21.56 | 37.02 |
| variance | 9.69 | 7.60 | 10.22 | 11.73 | 10.12 | 19.28 | 15.23 |

In this study, we evaluate all comparison models in automation evaluation. Among all comparison models, HuatuoGPT2-7B is unable to generate the TCM prescription. The generated content is the meaningless description about the patient's symptoms. Thus, we do not present HuatuoGPT2-7B's scores. The randomly selected generate screenshot can be found in **Appendix D**. As shown in Table 2, GPT-5, Deepseek-R1, Gemini-2.5-pro, Qwen3-235B-A22B-Thinking, and Med-Shicheng shows the consistent and great margin than other models. The Gemini-2.5-pro obtains the best score with the relatively low variance, which means stable prediction with the changing conditions of patients. Med-shicheng obtains the better score than Deepseek-R1 with higher variance, illustrating Med-shicheng fail to generate the similar TCM prescription with label on some cases. This can be attributed to the model's 1.5-billion-parameter, which limits its capacity for comprehensive data retention and induces a degree of knowledge forgetting. Despite this, Med-Shicheng attains a level of mean performance equivalent to Deepseek-R1, signifying that it can achieve predictive accuracy comparable to their general-purpose counterparts in the given task. The fact that this compact model was trainable under limited resource constraints, in contrast to the excessive parameter counts of large models, further attests to the efficacy of our proposed training framework in zero-shot learning setting. For Tianyi and Qwen2.5-1.5B-instruct, Qwen2.5-1.5B-instruct obtains better score than Tianyi but with the highest variances among all 7 models, which indicates Qwen2.5-1.5B-instruct might only generate very few cases with valid content. Tianyi can consistently generate some herbs along with Dr. Xiaohong Gu's theory to some extent.

## Discussion

In this work, we propose a standard and universal framework to model the specialized medical knowledge and skills of the distinguished physicians with the very lightweight LLM, i.e., Qwen2.5-1.5B-Base, called Med-shicheng. The comparable evaluation results with SOTA LLMs (e.g., Deepseek-R1, GPT-5, Gemini-pro-2.5) of both LLMs and human TCM doctors prove its effectiveness and efficiency on both data utilization and model performance improvement. The promising evaluation results show the potential of lightweight LLMs as the assistant of complicated tasks, limited computational and data resources, professional domains. In our evaluations, the general-purpose LLMs with extremely large-scale parameters show their capabilities on evaluating informative long medical response with the detailed designation of instructions. The Med-shicheng framework is not only applied in TCM, but also feasible for the high-effectiveness specialized medical knowledge and skills modeling in modern medicine. A comprehensive strategy and formation clinical diagnostic-and-therapeutic by modern medicine is presented in the **Appendix A.2**.

Besides, we treated the results of Human Evaluation as the ground truth to examine the feasibility of using ***LLM as a Judge*** in the medical field. Existing work [29,30] on LLM as a Judge has pointed out that when using LLMs as judges, statistically significant results often lack generalizability, as such generalizability only holds incidentally under specific "evaluated model + evaluation model" combinations. Therefore, the Jonathan Liu *et al.* [29] recommend using multiple LLMs as judges in the absence of ground truth or physician-annotated data. However, a limitation of their study is the that whether aggregating the results of multiple LLMs can truly mitigate the challenge of poor generalizability in LLM-based evaluation. Our work provides some clues and evidences for the issue remained in their study[29]. by comparing Automation Evaluation and Human Evaluation results, we noticed that although the aggregated evaluation results from multiple LLMs show a certain degree of consistency with the assessments of human physicians, notable differences remain at more detailed, individualized clinical levels. Presented in automation evaluation results, Med-shicheng performs comparably to current SOTA general-purpose LLMs, but outperforms them in Human Evaluation, which more accurately and professionally distinguishes model performance differences. These findings suggest that LLMs serving as judges require specialized training in the medical domain to achieve reliable judgment. Without adequate domain knowledge, LLMs often misinterpret subtle clinical meanings and requirements, leading to erroneous and harmful evaluations—even in settings where multiple LLMs are aggregated. This underscores the critical importance of domain-adapted training for enhancing the evaluation capability of LLMs as judges in specialized fields.

In light of the above analysis, we recommend that current medical LLM benchmarks exercise caution when relying solely on LLM even though the aggregated results from multiple LLMs as judges without ground truth or physician-annotated data. While aggregating multiple LLMs' evaluations may help reveal variance in the evaluated models, we argue that involving human physicians in model assessment is indispensable when corresponding ground truth is unavailable. Particularly in more complex clinical scenarios, merely aggregating multiple LLM judgments may conflate the limitations of the LLMs with the shortcomings of the evaluated models, potentially leading to misleading conclusions. Therefore, when adopting the **LLM-as-a-Judge paradigm** to evaluate domain-specific models, it is essential to first assess whether the selected LLMs possess sufficient domain-specific expertise. Otherwise, the evaluation results may reflect the knowledge limitations of the judging models rather than the true capabilities of the evaluated models.

## Conclusion

As a fundamental and persistent challenge in clinical medicine: how to disseminate and transmits of the personalized, holistic diagnosis and treatment knowledge systems possessed by distinguished physicians is discussed and studies for decades, where numerous studies are emerged trying to accomplish this goal. In our opinion, modeling expertise of a distinguished physicians is not a simple collection of facts or classification tasks but a sophisticated, internally consistent framework forged through decades of integrating foundational knowledge with reflective clinical practice. The traditional apprenticeship way of knowledge transfer is inherently limited, creating a scarcity of expertise, widening skill disparities among clinicians, and exacerbating healthcare inequities. The central challenge, therefore, lies in developing a standardized, scalable method to archive and replicate this nuanced, individualized clinical reasoning.

In response, we propose **Med-Shicheng**, a general methodology designed to systematically model and transmit the comprehensive knowledge systems of distinguished physicians into an agent, such as a LLM. The research rationale is built on the premise that to authentically emulate expert-level practice, an AI system must move beyond task-specific predictions and construct interconnected reasoning pathways that mirror the physician's holistic decision-making process under the guidance of target knowledge system of a distinguished physician. This requires an anthropomorphic modeling approach that captures the physician's comprehension of foundational knowledge, their reflective refinement of diagnostic principles, and their continuous reconciliation of theory with clinical practice. The Med-Shicheng framework operationalizes this through a structured, five-stage methodology. **First**, we construct a medical in-domain foundation model to establish a robust base of general medical knowledge. **Second**, the model learns regular TCM D&T strategies to grasp standard clinical protocols. **Third**, we employ reasoning-based reinforced fine-tuning to strengthen the model's clinical decision-making logic. The core innovation lies in the final two stages: **fourth**, the model undergoes unique medical knowledge fine-tuning using authoritative research articles and clinical case portfolios from target distinguished physicians, which encapsulate their characteristics clinical philosophy and reasoning logic; and **fifth**, a reasoning-based reinforced learning stage specifically targets the refinement of these unique D&T strategies. This multi-stage process effectively leverages both labeled and unlabeled data, organizing diverse resources—from systematic disease treatises to primary-source case evidence—into tailored tasks that collaboratively build a coherent model of the physician's expertise. A key feature of our approach is the successful integration of multiple, distinct physician knowledge systems into a single model without inducing negative interference, eliminating the need for labor-intensive individual models for each expert.

The extensive experiments shows that **Med-Shicheng is a novel and generalizable solution for the standardized transmission of personalized medical knowledge and skills**. It demonstrates a highly efficient methodology for integrating heterogeneous data sources to construct a holistic model of clinical expertise, achieving remarkable performance with a very small set of labeled samples and model scale that our implementation proves the significant potential of lightweight, cost-effective LLMs in specialized domains. By utilizing the compact Qwen2.5-1.5B-Base model within the Med-Shicheng framework, we achieved performance competitive with the massive 671-billion-parameter DeepSeek-R1 model, despite a 447-fold reduction in parameters. This breakthrough demonstrates that with a carefully designed, domain-specific training paradigm, small-scale models can excel at complex, multi-faceted clinical tasks, making advanced AI assistance feasible for deployment on limited computational resources, such as consumer-grade hardware, and opening new avenues for accessible, in-domain AI research. Besides, **this framework provides an effective strategy for the harmonious integration of diverse medical knowledge systems within a single model**. This capability is crucial for building comprehensive clinical AI assistants that can draw upon the synthesized wisdom of multiple experts without confounding their distinct methodologies. An applicable and comprehensive strategy for modern medicine clinical diagnosis and treatment is presented according to the Med-shicheng framework, illustrating its versatility. We also analyze the feasibility of taking **LLMs-as-Judges** paradigm, which is proven to be the preliminary stage on evaluating informative long content that solely relies on LLMs. In summary, Med-Shicheng represents a transformative step toward overcoming the critical bottleneck of clinical expertise transmission. By enabling the scalable, high-fidelity modeling of distinguished physicians' knowledge systems within a deployable, lightweight model, this work holds profound implications for augmenting the skills of frontline clinicians, democratizing access to specialized expertise, and ultimately, mitigating disparities in healthcare quality.

## Datasets

In this study, we developed a series of tailored datasets to support the multi-stage Med-Shicheng framework. All evaluation data for comparative models were independently collected. The types and volumes of the training datasets used are summarized in Table 3.

The datasets were constructed as follows: **First**, we performed incremental pre-training of a general large language model using the **TCMCorpus**—previously employed to train Tianyi—to adapt it to the TCM domain. TCMCorpus comprises a large-scale collection of TCM texts, including standardized textbooks, classical works and books, TCM specialized books, and clinical research papers. **Second**, to equip the model with routine D&T strategies and clinical reasoning skills, we integrated multiple TCM resources—clinical records, consultation transcripts, examination books, knowledge-style Q&A, and the TCMMMKG knowledge graph—into a unified instruction set named **TCM-general-instructions**. This formed and applied at Stage 2: TCM Regular D&T Strategy Learning. **Third**, to cultivate comprehensive D&T reasoning—covering etiology analysis, syndrome diagnosis, treatment principles, TCM prescription generation, explanation, outcome prediction, and medical advice—we designed Stage 3: Reasoning-based D&T Reinforced Fine-tuning. We annotated real-world clinical cases with full diagnostic reasoning pathways, resulting in the **TCM-reasoning dataset**. **Subsequently**, in Stage 4: Unique TCM Knowledge Fine-tuning, we introduced **TCM-unique-instructions**, an instruction dataset built from clinical cases of distinguished TCM physicians. This enables the model to internalize distinctive diagnostic philosophies and prescribing habits. **Finally**, in Stage 5: Reasoning-based Unique D&T Strategy Reinforced Learning, we expanded the master physicians' cases and filtered out inconsistent treatment plans. These were blended with original samples and formatted into the **TCM-unique-reasoning dataset**. We applied the KTO[31] algorithm for reinforcement learning, using both preferred and model-rejected samples generated during fine-tuning. **All electronic medical records were de-identified to protect patient privacy.**

Table 3. The datasets we built for the training in Med-Shicheng framework.

| Stage | Name | Dataset Type | Number of Samples | Total Tokens |
|---|---|---|---|---|
| Medical in-domain foundation model construction | TCMCorpus | the National planned textbooks for higher education in TCM, publications of different TCM schools of thoughts, content of TCM research articles | 1.65 million | 3.4 billion |
| TCM regular D&T strategy learning | TCM-general-instructions | TCM clinical records, TCM consultation record of doctor-patient, TCM examination QA Books, TCM knowledge graph | 0.3 million | 8 million |
| reasoning-based D&T reinforced fine-tuning | TCM-reasoning | Reasoning chains supplemented general clinical cases | 96 thousand | 2.17 million |
| unique TCM knowledge fine-tuning | TCM-unique-instructions | Knowledge-supplemented Clinical cases of target distinguished TCM practitioner | 0.5 million | 4.12 million |
| reasoning-based unique D&T strategy reinforced learning | TCM-unique-reasoning | Clinical cases of target distinguished TCM practitioner in KTO-format | 0.4 million | 9.35 million |

## TCMCorpus Dataset

**TCMCorpus** is a corpus encompassing a wide range of TCM medical and clinical texts, designed for TCM pre-training. It integrates a total of 1.65 million samples from three sources, amounting to 3.4 billion tokens. The specific sources are as follows: (1) **National Planned TCM Textbooks for Higher Education in China**: This includes 126 types of textbooks from the "14th Five-Year Plan" for higher education in TCM, covering disciplines such as TCM studies, Chinese materia medica, acupuncture and moxibustion, tuina, integrated Chinese and Western clinical medicine, and nursing.

These textbooks were compiled by over 3,500 experts from more than 30 higher education institutions across China, representing authoritative achievements in TCM higher education. (2) **Publications from Various TCM Schools**: This collection includes relevant publications from ten major TCM schools, namely the Classical Medicine School, Classical Formula School, Cold Damage School, Hejian School, Attack and Elimination School, Nourishing Yin School, Yishui School, Warm Tonification School, Warm Disease School, and Syncretic School, with a sample size of 1.5 million entries. (3) **TCM Research Papers**: various clinical research and case studies in TCM by renowned TCM practitioner were collected, including the TCM doctors we are aiming to study: the professors/doctors Gu Xiaohong, Du Huaitang, Wang Fengchun, Zhao Shaoqin, and Qin Bowei, totaling 99,000 samples. The corpus has undergone data preprocessing, including the removal of special markers and characters, and has been structurally processed to ensure data quality.

### TCM-General-Instructions Dataset

To enable the model to grasp the fundamental diagnostic and treatment logic of TCM, we incorporated high-quality Chain-of-Thought (CoT) [32] reasoning patterns to construct the training samples. This dataset contains 302,000 samples, with a total of 8 million tokens. It primarily consists of the following content: **(1) TCM Clinical Records:** 30,000 samples sourced from real-world TCM electronic medical records, covering various diseases across different systems, such as respiratory febrile diseases and gastrointestinal disorders. **(2) TCM Doctor-Patient Consultation Records:** 2,000 samples derived from audio recordings of actual consultations conducted by renowned TCM practitioners. These include doctor-patient dialogues, analysis of disease etiology and mechanisms, syndrome diagnosis, distinctive pattern differentiation thinking, herbal prescriptions, and formula explanations. **(3) TCM Knowledge Books:** In addition to the aforementioned data types which inherently possess CoT characteristics, we reconstructed content from a category of TCM knowledge books, such as "xxx *Syndrome Compendiums*" and "xxx *Prescription*". These books typically contain entries like descriptions of syndromes or prescription, analysis of syndrome causes, indications, treatment principles, and common formula compositions. Following the logic of TCM clinical practice, we designed a set of reasoning chains that organize these entries according to specific logical sequences, thereby generating a batch of standardized TCM D&T CoT samples with unified formation and writing. Detailed sample design methods and examples can be found in **Appendix A.1**. By introducing the CoT structure into the TCM instruction samples, the model can systematically learn how to perform reasoning analysis based on this knowledge and clinical symptoms, laying a foundation for subsequent tasks.

### TCM-Reasoning Dataset

At this phase, to enhance the model's generalization capability and flexibility in target 7 TCM tasks, we performed output diversification on all samples containing CoT reasoning logic. This involved using the model from the previous phase to generate multiple outputs for the same input. Through the rejection sampling strategy, outputs that were highly consistent with the input question, TCM theory, and TCM logic were retained, while erroneous outputs were filtered out. This approach of model-based diversification strengthens the model's ability to generate varied and generalized correct outputs, while ensuring adherence to the TCM chain of thought. Subsequently, we applied the Group Relative Policy Optimization (GRPO) reinforcement learning method to guide the model in exploring personalized reasoning paths, learning from errors, and gradually converging to correct logical processes. The specific steps are as follows: (1) The model trained with GRPO was used to perform inference on the "Instruction" fields of the CoT samples from the TCM-general-instructions dataset via the vLLM framework, generating 10 responses per sample. (2) Using a Best-of-N approach, Deepseek-V3.2 was employed to score the responses with the specific prompt. (3) Samples with scores greater than 8.5 were selected to form the final augmented dataset, which contains 96,000 samples and totals 2.18 million tokens.

## TCM-Unique-Instruction Dataset

The objective of constructing this dataset is to enable the model to comprehensively learn the clinical diagnostic-and-therapeutic characteristics, experiences, and theories of targeted renowned TCM practitioners, building upon its mastery of conventional TCM diagnostic and therapeutic logic and capabilities, which originate from the actual clinical cases of target famous TCM doctors. Based on their published publications, we summarized their clinical characteristics and designed logical reasoning samples that reflect their distinctive styles according to the CoT structure and TCM diagnostic logic. Besides the CoT structure, we also drew inspiration from the Retrieval-Augmented Generation (RAG)[33] approach during model instruction tuning, strengthening model's ability to learn the application of these characteristic theories in clinical practice. We constructed expert knowledge bases for doctors Gu Xiaohong, Du Huaitang, Wang Fengchun, Zhao Shaoqin, and Qin Bowei. Each knowledge base consists of a set of text sequences, each containing 512 tokens after tokenization. For each clinical case from these renowned doctors, we performed content matching against the knowledge base. The top-3 matching expert knowledge chunks were concatenated after the clinical case information in a specific format as model input, namely: "【Knowledge Base 1】 content1\n 【Knowledge Base 2】 content2\n 【Knowledge Base 3】 content3." Additionally, at this stage, we also incorporated non-reasoning samples, specifically knowledge-type samples, including: **TCM Knowledge Q&A Data:** 100,000 samples sourced from real TCM examination questions, including the TCM Practicing Physician Qualification Exam and TCM postgraduate entrance exercises. **TCM Knowledge Graph:** 20,000 samples derived from the TCMMMKG (TCM Multi-modal Knowledge Graph) constructed in our previous research. The purpose of adding these samples is to enable the model to perform other conventional tasks, such as knowledge question answering, while maintaining its logical reasoning capabilities based on TCM knowledge.

The specific steps for the sample construction process are as follows: (1) Collect published books and papers of the target renowned TCM doctors and extract their main textual content. (2) Use the model trained during the GRPO phase to tokenize the knowledge base, generating tokenized sequences. (3) Chunk the knowledge base into sizes of 512 tokens. During chunking, segmentation is performed at punctuation marks of complete sentences to ensure the semantic integrity of each expert knowledge chunk. (4) Use the Gte-Qwen1.5-7B Chinese text embedding model to calculate the similarity between the chunked texts and the cases of the renowned doctors. Select the top three scoring knowledge chunks and populate them into the "input" field of the dataset.

## TCM-Unique-Reasoning Dataset

To steer the model's preference towards the distinctive diagnostic and treatment experiences of the target renowned TCM practitioners, we constructed a new dataset containing "true" or "false" labels, based on the **TCM-Shicheng-Instruction dataset**, which contains 381,000 samples, totaling 9.353 million tokens. During the construction of this dataset: (1) The original output entries from the source cases were directly assigned a "true" label. (2) We first constructed a binary classification model from a previous dataset training phase to serve as a scoring model. This model evaluates the semantic similarity between the outputs generated by a model for the cases of the target renowned doctors in the TCM-Shicheng-Instruction dataset and the original case outputs. Outputs with a semantic similarity exceeding 90% were labeled "true", while those with a similarity below 60% were labeled "false". (3) Finally, the optimization for multiple preference objectives was unified through KTO (Knowledge Transfer Optimization). Compared to traditional reinforcement learning, KTO demonstrates greater robustness to noisy data and can effectively handle data imbalance scenarios (e.g., where 90% of the samples are high-quality).

## Model overview

In this study, we propose Med-Shicheng, a universal framework for building LLM capable of systematically learning and inheriting the distinctive diagnostic and treatment knowledge systems of human physicians. The framework was validated on the task of preserving and transmitting the clinical experience of renowned distinctive doctors in TCM. Med-Shicheng comprises five training stages, utilizing Qwen-2.5-1.5B-Base as the base model. Through diversified TCM task design, it adapts a general LLM to the medical domain, transforming it into a D&T "expert" to achieve optimal performance on target tasks. By applying into TCM setting, these stages include: (1) **Medical In-Domain Foundation Model Construction**: This stage transforms a general-purpose base LLM into a foundational LLM specialized in the TCM domain. (2) **Regular Diagnosis & Treatment Strategy Learning**: The basic TCM LLM learns standard diagnostic and treatment logic within TCM. (3) **Reasoning-Based D&T Reinforced Fine-Tuning**: The TCM LLM enhances its reasoning and deduction capabilities, building upon the standard TCM D&T logic. (4) **Unique Medical Knowledge Fine-Tuning**: The TCM LLM, now equipped with standard TCM reasoning logic, learns the knowledge systems of distinguished TCM practitioners. (5) **Reasoning-Based Unique D&T Strategy Reinforced Learning**: The TCM LLM refines its diagnostic and treatment logic to emulate the approaches of master TCM practitioners.

### Medical in-domain foundation model construction

We selected the Qwen-2.5 series as the foundation for building a general model in the field of TCM. The training corpus of the Qwen-2.5 series contains 18 trillion general tokens, and among them, qwen2.5-1.5B-Base is currently the best open-source Chinese language model with a size that can operate efficiently on extremely limited computational resources. It possesses strong Chinese comprehension capabilities and a certain level of foundational knowledge in TCM, making it highly suitable as a base model for building in the TCM domain. At this stage, our objective is to significantly enhance the model's mastery of TCM knowledge while preserving the general capabilities of the qwen2.5-1.5B-Base model to the greatest extent possible. To achieve this, we conduct large-scale unsupervised autoregressive training based on the TCMCorpus dataset, using the loss function of a causal language model as the training objective. Let $\mathcal{D}_{TCMCorpus} = \{x_1, x_2, \ldots, x_N\}$ represent the collection of TCM samples, where each $x_i = [x_i^1, x_i^2, \ldots \ldots, x_i^{2048}]$ consists of tokenized semantic units of TCM text. The training objective at this stage is to minimize the negative log-likelihood function, as show in equation (1). The model trained in this phase is named **TCM-SC-Base**.

$$\mathcal{L} = -\sum_i^N \sum_j^{2048} \log P(x_i^j \mid x_i^1, x_i^2, \ldots \ldots, x_i^{j-1}; \theta) \qquad (1)$$

### TCM regular D&T strategy learning

In the previous phase, we completed incremental training of the general large model on medical knowledge, enhancing its mastery of TCM domain knowledge to be more profound and comprehensive. Then, applying TCM-general-instructions dataset, we proceed to stimulate the model's reasoning logic for routine TCM diagnostic-and-therapeutic, as well as its ability to answer questions on general TCM-related knowledge, based on TCM-Base. In this stage, we adopt a supervised learning approach and conduct two epochs of training on all samples in the TCM-general-instructions dataset. The training objective is to minimize the cross-entropy between the model's response and the supervised labels, as shown in equation (2). The model obtained upon completion of this training phase is named **TCM-SC-normal-instruct**.

$$\mathcal{H}(response, label) = -\sum_i label(i) * \log response(i) \qquad (2)$$

### Reasoning-based D&T reinforced fine-tuning

The TCM-SC-normal-instruct model gained the capability to generate a complete set of diagnostic-and-therapeutic plans as case outputs based on the learned TCM diagnostic knowledge and the reasoning logic of routine clinical cases, utilizing relevant patient case information. However, due to the limitation of training methodology, the D&T plans generated by TCM-SC-normal-instruct suffer from a significant limitation: the content of the plans tends to be overly rigid, lacking sufficient flexibility in personalized analysis of cases and the ability for contextual association. To address this issue and enhance the logical reasoning flexibility and innovation of TCM-SC-normal-instruct, we employ the TCM-reasoning dataset and adopt the GRPO algorithm in a reinforcement learning approach to improve its adaptability in logical reasoning. The model resulting from this phase of training is designated as **TCM-SC-normal-reasoning**.

### Unique Medical knowledge fine-tuning

After completing the aforementioned stages, the TCM-SC-normal-reasoning has acquired relatively flexible reasoning capabilities for TCM case diagnostic-and-therapeutic. It can dynamically integrate general TCM knowledge it learned with patients' clinical conditions to deliver personalized diagnostic and therapeutic plans, including etiology analysis, syndrome diagnosis, treatment principle recommendations, TCM prescription generation, TCM prescription explanation, post-medication symptom prediction, herbal modification, and precautions. To enable TCM-SC-normal-reasoning to learn and master the distinctive diagnostic and treatment methods or unique therapeutic concepts and logic of multiple target distinguished TCM physicians, we conduct a refined fine-tuning process using the TCM-Shicheng-instruction dataset. This step focuses on specialized TCM knowledge and diagnostic philosophy. Unlike Stage 2, we preserve both knowledge-type samples and conventional TCM case reasoning samples alongside the newly constructed reasoning samples of target master physicians at this stage. This ensures that TCM-SC-normal-reasoning retains its general clinical reasoning ability and knowledge-based question-answering capacity in TCM. The model obtained upon completion of this training phase is named **TCM-SC-unique-instruct**.

### Reasoning-based unique D&T strategy reinforced learning

After the learning of the distinctive TCM knowledge and diagnostic logic of the target TCM physicians, we further consolidate TCM-SC-unique-instruct's alignment with the characteristic knowledge systems and diagnostic reasoning preferences of these distinguished TCM physicians, while enhancing its flexible diagnostic reasoning capabilities. To achieve this, we employ the independently constructed TCM-unique-reasoning dataset and utilize the KTO reinforcement learning algorithm to strengthen the aforementioned abilities of TCM-SC-unique-instruct. The core objective functions for this stage are shown in Equations (3) and (4) below:

1) reference label, $z_{ref}$, is mathematically defined as the expectation of payoffs when the optimal strategy is implemented

$$z_{ref} = E_{x' \sim D}[\beta KL(\pi_\theta(y'|x')||(\pi_{ref}(y'|x'))] \qquad (3)$$

2) Value function, $v(z)$, describes the subjective value of gains or losses relative to a reference point (preferred response):

$$v_{KTO}(z) = \begin{cases} \sigma\big(r_{KTO}(x,y) - z_{ref}\big), if \ y{\sim}y_{desirable} \\ \sigma\big(z_{ref} - r_{KTO}(x,y)\big), if \ y{\sim}y_{undesirable} \end{cases} \tag{4}$$

where，$z_{ref}$ is Kullback-Leibler divergence and $r_{KTO}$ is the reward function used in the RL algorithm.

## Acknowledgements

The authors would like to thank physicians (T. Mao, F. Liu, X. Li) from the Dongfang Hospital, Beijing University of Chinese Medicine, physician (C. Wu) from Shenzhen Hospital, Beijing University of Chinese Medicine, physician (W. Wu) from Hubei Provincial Traditional Chinese Medicine Hospital, physician (Y. Wang) from Shunyi Hospital, Beijing Hospital of Traditional Chinese Medicine, physician (Y.S. Zhao, G. Liu, Y. He, C. Bai, X.H. Gu) from department of Warm Disease Studies, Beijing University of Chinese Medicine, physician (Y. Li, and H.T. Du) from Dongzhimen Hospital, Beijing University of Chinese Medicine, physician (B.C. Wang) from department of Warm Disease Studies, Nanjing University of Chinese Medicine, physician (J. Wang) Wangjing Hospital, China Academy of Chinese Medical Sciences, physician (Y.R. Kong) from Beijing Hospital of Traditional Chinese Medicine (Huairou Hospital), physician (Y.F. Chen) from Xiamen Medical College, (X. Yang) from Fangshan Hospital Beijing University of Chinese Medicine.

## Author contributions

# Appendix

# A. Chain of Thought strategy for Clinical Diagnosis and Treatment

### A.1 Chain of Thought strategy for TCM

We design a unified strategy that widely used in TCM clinical diagnosis and treatment process. It contains the following steps: (1) etiology and pathogenesis analysis in TCM, (2) syndrome differentiation, (3) treatment principle, (4) TCM prescription generation, (5) prescription explanation, (6) application of distinguished or specialized differentiation and treatment theory (optional), (7) modification of herbs based on symptom changes, and (8) medical advice and precautions. All distinguished TCM doctors' original valid cases and the general TCM clinical cases are expended COT process in both reasoning pattern and formal output pattern along with the abovementioned 8 steps. We use template <think></think>\n<output></output> to form the output that Med-shicheng should learn. The format we expend the COT reasoning part and formal part is following:

<think>

According to the given patient's symptoms and medical history, I need to analysis patient's etiology and pathogenesis analysis in TCM. First, …… Then, the syndrome differentiation should be …… The treatment principle will be applied with …… Next, the candidate herbs are …… Thus, the prescription should be …… The concrete herbs are …… After that I should analysis the functions of these medicines, …… The process of syndrome differentiation, treatment principle making and TCM prescription is applied according to …… (if needed) When patient take these herbs, if one symptom goes worse, then some herbs should replace by …… Some herbs should removed or reduce it dose if some symptoms are relieved. During taking this TCM prescription, patient should take care of the following precautions: ……

</think>\n<output>

**Etiology and Pathogenesis Analysis (in TCM)**

……

**Syndrome Differentiation**

……

**Treatment Principle**

……

**TCM Prescription**

……

**Prescription Explanation**

……

**Application of Distinguished or Specialized Differentiation and Treatment theory**

……

**Modification of Herbs Based on Symptom Changes**

……

**Medical Advice and Precautions**

……

</output>

**A.2 Chain of Thought strategy for modern medicine clinics**

A similar diagnosis and treatment strategy and formation for the modern medicine clinic can also be constructed accordingly:

**1. Conceptual Correspondence in Modern Medicine**

The core objective is to enable the model to learn the underlying reasoning logic across complex cases, rather than simply memorizing static mappings between symptoms, patterns, and prescriptions. Thus, in modern medicine, this can be correspond to the following components:

1. **Expert-Level Disease Concepts and Theoretical Systems**

   This corresponds to how leading clinical experts in a given field conceptualize disease, including:

   o Their particular understanding of pathophysiology (e.g., an expert in heart failure emphasizing the integrated "heart–kidney–metabolic" axis).

   o Their customized subtyping systems (e.g., refined stratification of disease phenotypes or risk categories).

   o Their self-developed clinical pathways or diagnostic–therapeutic schemas (e.g., decision trees and structured workflows).

2. **Distinctive Diagnostic and Therapeutic Techniques**

   This corresponds to:

   o **Diagnostic level**: unique approaches to differential diagnosis, characteristic weighting and prioritization of clinical signs, laboratory values, and imaging findings, and specific triggers for suspecting rare or atypical diseases.

- o **Therapeutic level**: distinctive drug selection and titration strategies, preferred combinations of therapies, decisions on timing and indications for escalation to interventional or surgical procedures, and nuances in perioperative or critical care management.

3. **Monographs, Reviews, and Case-Based Theoretical Expositions**
   This includes:
   - o Expert-authored monographs and specialty textbooks, contributions to clinical guidelines, and high-impact narrative or systematic reviews.
   - o Records of ward rounds, multidisciplinary discussions, and case conferences.
   - o Case reports and case series in which complex or rare cases are analyzed in depth, including rationale for each diagnostic and therapeutic decision.

4. **Real Clinical Cases and Systematic Case Analyses**
   This comprises:
   - o High-quality electronic health records with longitudinal follow-up.
   - o Typical and atypical cases from expert outpatient clinics.
   - o Multidisciplinary team (MDT) consultation records and critical care logs.
   - o Documentation of adverse outcomes and failed treatments, together with reflective analyses, which are crucial for training robust clinical reasoning.

Taken together, what needs to be "inherited" in modern medicine is not merely the correctness of individual diagnoses or treatments in isolation, but an expert's integrated disease model and decision-making strategy. The goal is to capture the full reasoning process that underlies the continuum from problem recognition, through diagnosis and treatment, to explanation and follow-up adjustment.

**2. A Seven-Step Clinical Reasoning Paradigm in Modern Medicine**
The TCM application organizes reasoning into seven interconnected tasks (e.g., aetiology–pathogenesis analysis, pattern differentiation, therapeutic method selection, prescription design, formula explanation, reflection of the master's theory, and adjustment according to symptom evolution). In modern medicine, a similar process can be formulated as follows:

1. **Etiology and Pathophysiology Analysis**
   This step corresponds to TCM aetiology–pathogenesis analysis and focuses on:
   - o Identifying the primary etiological factors (infection, autoimmunity, genetic defects, metabolic disorders, environmental exposures, lifestyle factors, etc.).
   - o Analyzing the key pathophysiological pathways (e.g., renin–angiotensin–aldosterone system activation, inflammatory cascades, endothelial dysfunction, disordered coagulation).
   - o Explaining the relationships between clinical findings (signs, symptoms, tests) and the underlying mechanisms.

2. **Diagnosis and Differential Diagnosis**
   This step parallels pattern differentiation and includes:
   - o Defining the working diagnosis, supported by integrated clinical reasoning.
   - o Enumerating and prioritizing differential diagnoses, specifying supporting and refuting evidence for each candidate disease.
   - o Explicitly linking diagnostic hypotheses to the data sources: medical history, physical examination, imaging, laboratory parameters, genetic or molecular tests, and functional assessments.

3. **Therapeutic Goals and Overall Strategy**

   This corresponds to determining the "therapeutic principle" in TCM:

   o Defining short-term goals (e.g., symptom relief, hemodynamic stabilization, control of infection).

   o Defining mid- and long-term goals (e.g., improving survival, preserving organ function, preventing complications and readmissions, enhancing quality of life).

   o Selecting the overarching treatment pathway:

   ▪ Predominantly pharmacological versus early interventional or surgical approaches.

   ▪ Conservatively stepwise escalation versus proactive, early intensive treatment.

   ▪ Curative intent versus supportive, rehabilitative, or palliative focus.

4. **Design of the Concrete Treatment Regimen**

   This step corresponds to prescription design in TCM:

   o **Drug therapy**:

   ▪ Selecting first-line and second-line medications and their combinations.

   ▪ Determining dosages, titration schedules, routes of administration, and duration of therapy.

   ▪ Identifying and managing potential drug–drug interactions, contraindications, and patient-specific risks.

   o **Non-pharmacological interventions**:

   ▪ Planning surgical or interventional procedures and perioperative care.

   ▪ Determining required monitoring intensity (e.g., general ward vs. high-dependency unit vs. ICU).

   ▪ Designing lifestyle modifications, rehabilitation programs, and psychological or social support.

5. **Mechanistic and Evidence-Based Justification**

   This corresponds to the detailed "formula explanation" in TCM:

   o Providing a mechanistic rationale for each major intervention (e.g., how a given drug modifies neurohormonal activation, inflammatory responses, ventricular remodeling, or metabolic pathways).

   o Citing the key evidence base that supports the decision (e.g., landmark randomized controlled trials, meta-analyses, and guideline recommendations with strength-of-evidence grading).

   o Explaining why apparently plausible alternatives are not recommended, whether due to insufficient evidence, unfavorable risk–benefit balance, or patient-specific contraindications.

6. **Expression of Individualized Expert Clinical Philosophy**

   This step parallels the manifestation of a renowned TCM physician's distinctive diagnostic and therapeutic philosophy:

   o Demonstrating how, within the same guideline framework, an expert re-calibrates risk–benefit judgments (e.g., opting for earlier combination therapy, or conversely, adhering to a more conservative titration in complex comorbid profiles).

- o Showing characteristic strategies for special populations (e.g., the very elderly, pregnant patients, individuals with rare comorbidities or extreme frailty).
- o Encoding the expert's tacit rules-of-thumb and heuristics, such as specific patterns of subtle clinical or laboratory changes that trigger pre-emptive regimen adjustments.

7. **Prediction of Disease Trajectory and Dynamic Treatment Adaptation**

   This corresponds to anticipating symptom evolution and adjusting the prescription or "adding/removing herbs" in TCM:

   - o Predicting short- and long-term trajectories, including risks of complications (e.g., decompensated heart failure, progressive renal dysfunction, thromboembolic or hemorrhagic events).
   - o Planning follow-up and re-evaluation schedules, specifying which parameters should be monitored (imaging, laboratory markers, functional scores, patient-reported outcomes).
   - o Defining explicit if–then adaptation rules, such as:
     - ▪ **If** specific thresholds or trends in biomarkers, imaging findings, or clinical status are observed, **then** escalate, de-escalate, or switch therapies accordingly.
   - o Integrating unexpected responses or adverse events into a continuous learning loop, refining future decisions for similar patient profiles.

This paradigm provides a structurally parallel, modern-medical formulation of the original TCM-inspired framework and can serve as a conceptual foundation for training and evaluating domain-specific clinical reasoning models with Med-shicheng framework.

# B. The Human Expert Evaluation Standards and Evaluation Prompts of Deepseek-V3.2 and GPT-5

## B.1. Evaluation Prompts for Deepseek-V3.2 and GPT-5 as the judges of automation evaluation

We design the detailed evaluation instruction prompt for Deepseek-V3.2 and GPT-5 based on the evaluation framework in section B.1.

➢ **Evaluation Instruction prompt for Deepseek-V3.2.**

**SYSTEM PROMPT**

You are a professional TCM evaluation expert. Your task is to compare the TCM diagnostic content generated by an AI model against a standard answer (label) and score/analyze the following five distinct Items.

Please output the evaluation results strictly in the following JSON SCHEMA:

```
{
   "Completeness": {
      "score": 0,
      "Number of Items Actually Answered ": 0,
      "Total Number of Items Requiring Responses ": 5,
      "Missing Item": []
   },
   "Analysis of Etiology and Pathogenesis": {
      "score": 0,
      "Recognition of Etiology": 0,
      "Description of Pathogenesis": 0,
      "Logical Coherence ": 0
   },
   "Syndrome Differentiation": {
      "score": 0,
      "Accuracy of Syndrome": 0,
      "Disease Location and Nature ": 0
   },
   "Treatment Principle": {
      "score": 0,
      "Accuracy of Treatment Principle": 0,
      " Specificity of Treatment Method ": 0,
      " Application of Specialized Methods ": 0
   },

   "TCM Prescription": {
      "score": 0,
      " Medicinal Match Score ": 0,
      "Number of matched herbs": 0,
      "Number of Herbs in Label Prescription": 0,
      "Number of Herbs in Model-Generated Prescription": 0,
```

```
    "The List of Overlapped Herbs in both TCM Prescriptions ": [],
    "Matching rates": "0%"
  },
  "Distinguished Theory application": {
    "score": 0,
    "Accuracy of Academic Thought": 0,
    "Pervasiveness of Thought": 0,
    "Completeness of Elaboration": 0
  },
  "Total Score": 0,
  "Maximum Score": 55
}
```

**Detailed Evaluation Criteria:**

**0. Response Completeness (Basic points, Max 5 points)**

- **Calculation:** Score = (Number of Items actually answered / 5) × 5
- All 5 Items answered: 5 points
- 4 Items answered: 4 points
- 3 Items answered: 3 points
- 2 Items answered: 2 points
- 1 Item answered: 1 point
- No valid response or completely unanswered: 0 points

**Judgment Standard:**

- **"Valid Response" Definition:** The Item contains substantive content, not blank, "unknown," "cannot answer," or other invalid expressions.
- If an Item has only 1-2 sentences but contains substantive content, it is considered a valid response, but quality differences will be reflected in the specific scoring for that Item.

**1. Etiology and Pathogenesis Analysis (Max 10 points)**

- **1.1 Accuracy of Etiology Identification (4 points)**
  - o 4: Accurately identifies all main pathogenic factors; analysis is comprehensive and conforms to TCM theory.
  - o 3: Identifies main etiology (>80%), with only minor factors omitted.
  - o 2: Identifies some etiology (50%-80%), with certain omissions or slightly inaccurate.
  - o 1: Identifies only a few etiological factors (<50%), or contains obvious errors.
  - o 0: Completely incorrect or unanswered.

- **1.2 Completeness of Pathogenesis Elaboration (4 points)**
  - o 4: Fully elaborates the pathological progression, including the nature of pathogenic qi, disease location, disease tendency, etc.
  - o 3: Relatively complete elaboration, covering most core elements (>80%).
  - o 2: Basically reasonable but incomplete elaboration (50%-80%).
  - o 1: Incomplete elaboration (<50%).
  - o 0: Completely incorrect or unanswered.

- **1.3 Logical Coherence (2 points)**
  - o 2: Clear causal relationship from Etiology → Pathogenesis → Symptoms.

- o 1: Basically coherent, but with some logical jumps in parts.
- o 0: Illogical or self-contradictory.
- **Total Score Calculation:** Etiology Identification + Pathogenesis Elaboration + Logical Coherence

**2. Syndrome Differentiation (Max 10 points)**

- **2.1 Accuracy of Syndrome Diagnosis (6 points)**
  - o 6: Syndrome name fully conforms to TCM standards, consistent with the standard answer or uses an equivalent syndrome name.
  - o 5: Syndrome diagnosis is accurate, wording differs slightly but core meaning is the same (e.g., "Liver Depression and Spleen Deficiency" vs. "Liver-Spleen Disharmony").
  - o 4: Syndrome is basically accurate, covers the main Pathogenesis, but might omit secondary syndromes or wording is not precise enough.
  - o 3: Syndrome is partially accurate, captures some core elements but has certain deviations (e.g., correct only in disease location OR nature).
  - o 2: Syndrome diagnosis has significant deviation, but still has some relevance to the condition.
  - o 1: Syndrome is wrong, but disease location OR nature is partially correct.
  - o 0: Completely incorrect or unanswered.
- **2.2 Clarity of Disease Location and Nature (4 points)**
  - o 4: Clearly and accurately specifies disease location and nature.
  - o 3: Disease location and nature are basically clear, with 1-2 minor inaccuracies in wording.
  - o 2: Either disease location or nature is clear, the other is vague or deviated.
  - o 1: Expression of disease location and nature is vague or only partially correct.
  - o 0: Completely incorrect or not specified.
- **Total Score Calculation:** Syndrome Accuracy + Clarity of Location/Nature

**3. Treatment Principles (Max 10 points)**

- **3.1 Accuracy of Treatment Principle (5 points)**
  - o 5: The overarching treatment method is completely correct and fully corresponds to the syndrome.
  - o 4: Treatment principle is accurate, wording differs slightly but essence is the same.
  - o 3: Treatment principle is basically correct, but might omit secondary principles.
  - o 2: Treatment principle is partially correct, but has certain deviations.
  - o 1: Treatment principle has major deviations.
  - o 0: Treatment principle is wrong or unanswered.
- **3.2 Specificity of Treatment Method (3 points)**
  - o 3: Treatment method highly aligns with the etiology, Pathogenesis, and syndrome differentiation result.
  - o 2: Treatment method is relatively specific, but might lack consideration for minor aspects.
  - o 1: Treatment method has some specificity, but is not precise enough.
  - o 0: Treatment method lacks specificity or unanswered.
- **3.3 Application of Characteristic Methods (2 points)**

- o 2: Clearly embodies the physician's characteristic treatment ideas, applied appropriately.
- o 1: Somewhat embodied but not prominent.
- o 0: Not embodied.
- **Total Score Calculation:** Principle Accuracy + Method Specificity + Characteristic Methods

## 4. TCM Prescription (Max 10 points)

- **4.1 Medicinal Match Score (9 points)**
  - o **Calculation:** Score = (Number of identical medicinals / Total number of medicinals in standard answer) × 9
  - o **Matching Rule:** Extract all Chinese medicinal names from the standard answer and the model's generation. Count the number of identical medicinals.
  - o **Alias Handling:** Medicinals with the same efficacy but different names are considered identical. Medicinals with similar names but different efficacy are not considered identical. Medicinals from the same source but different processing methods are considered identical.

- **4.2 Formula Composition Logic (1 point)**
  - o 1: Overall formula composition is reasonable, conforms to the principles of Sovereign, Minister, Assistant, Envoy (Jun Chen Zuo Shi).
  - o 0.5: Composition is basically reasonable, but has minor flaws.
  - o 0: Contains incompatibility contraindications or clearly unreasonable combination of traditional Chinese medicines.

  **Example:**

  - Standard Answer: Huang Qin 10g, Huang Lian 10g, Jin Yin Hua 10g, Lian Qiao 10g, Tao Ren 10g, Hong Hua 10g, Dang Gui 10g, Chuan Xiong 10g, Chi Shao 10g, Gui Zhi 10g, Zhi Qiao 10g, Gan Cao 10g (Total: 12 medicinals)
  - Model Answer: Chai Hu 15g, Huang Qin 10g, Gui Zhi 9g, Bai Shao 12g, Ge Gen 20g, Huang Lian 6g, Tian Hua Fen 15g, Mu Dan Pi 10g, Chi Shao 10g, Sheng Jiang 3 slices, Da Zao 5 pieces, Gan Cao 6g (Total: 12 medicinals)
  - Identical Medicinals: Huang Qin, Huang Lian, Gui Zhi, Chi Shao, Gan Cao (5 medicinals)
  - Medicinal Match Score: 5/12 × 9 = 3.75 points
  - Composition Logic: Reasonable, 1 point
  - Total Score: 3.75 + 1 = 4.75 points
  - **Total Score Calculation:** Medicinal Match Score + Composition Logic

## 5. Application of Specialized TCM knowledge System (Max 10 points)

- **5.1 Accuracy of Academic Thought (5 points)**
  - o 5: Accurately embodies the core academic thinking of the specific physician.
  - o 4: Fairly accurately embodies the main academic thoughts.
  - o 3: Basically embodies the academic thought.
  - o 2: Partially embodies the academic thought.
  - o 1: Embodiment of academic thought is not obvious.
  - o 0: Not embodied or misunderstood.

- **5.2 Pervasiveness of Thought (3 points)**

- o 3: Academic thought permeates the entire process: etiology/Pathogenesis, syndrome differentiation, treatment method, formula/medicinals.
- o 2: Academic thought is reflected in multiple stages.
- o 1: Academic thought is only reflected in partial stages.
- o 0: Academic thought is not pervasive.
- **5.3 Completeness of Elaboration (2 points)**
  - o 2: Elaboration of the characteristic thought is clear and complete.
  - o 1: Elaboration is basically clear, but not complete enough.
  - o 0: Elaboration is unclear or absent.
- **Total Score Calculation:** Academic Thought Accuracy + Thought Pervasiveness + Elaboration Completeness

**Important Notes:**

1. **Objectivity and Fairness:** Scoring must be based on TCM theory and clinical practice, avoiding subjective bias.
2. **Response Completeness Priority:** First, evaluate whether the model has answered all Items completely. This is the basic scoring item.
3. **Medicinal Matching Rules:**
   - o Disregard differences in dosage, processing method, place of origin, etc.
   - o For alias medicinals, judge equivalence based on formula indication and efficacy.
   - o For medicinals with the same name but different sources (e.g., Nan Sha Shen / Bei Sha Shen), judge based on efficacy.
4. **Scoring Anchor:** Strictly adhere to the descriptions corresponding to each score. Each dimension has clear scoring criteria.
5. **Handling Missing Items:**
   - o If an Item is missing from the standard answer, that Item is not scored, and the total score is adjusted accordingly.
   - o If the model did not answer an Item, that Item scores 0 points, but it must be marked in the "Missing Items".
6. If any or all of the five Items in JSON SCHEMA are null/empty, directly assign a score of 0 for the corresponding item.
7. **Total Score Calculation:** Max Score = Response Completeness (5 points) + Total of five Items (50 points) = 55 points.

**USER PROMPT**

Please evaluate the model-generated content with the label based on the instructions provided in **SYSTEM PROMPT**, and output the evaluations according the provided JSON SCHEMA:

Label: {label_content}

Model response: {model_response}

Output scores:

The Deepseek-V3.2 will be fed with the instructions both SYSTME PROMT and USER PROMPT as input, of which contains the label content and model response in the variables of '*label_content*' and '*model_response*', respectively. Then, we can obtain the evaluation scores in the format that we defined in JSON SCHEMA.

➢ **Evaluation Instruction prompt for GPT-5.**

The instruction prompts format and evaluation dimensions detailed criteria for GPT-5 are identical to those in Deepseek-V3.2, so omitted for brevity in this optimized version; refer to above for details. We provide the JSON SCHEMA here for GPT-5. The JSON SCHEMA for GPT-5 evaluations is:

EVALUATION_JSON_SCHEMA: Dict[str, Any] = {

    "name": "tcm_evaluation",

    "schema": {

        "type": "object",

        "properties": {

            **"Response Completeness"**: {

                "type": "object",

                "properties": {

                    "score": {"type": "number"},

                    "Number of Items Actually Answered": {"type": "number"},

                    "Total Number of Items Requiring Responses": {"type": "number"},

                    "Missing Item": {"type": "array", "items": {"type": "string"}}

                },

                "required": ["score", " Number of Items Actually Answered ", " Total Number of Items Requiring Responses ", "Missing Item"],

                "Additional Properties": False,

            },

            " Analysis of Etiology and Pathogenesis": {

                "type": "object",

                "properties": {

                    "score": {"type": "number"},

                      "Recognition of Etiology": {"type": "number"},

                    "Description of Pathogenesis": {"type": "number"},

                    "Logical Coherence": {"type": "number"}

                },

                "required": ["score ", " Recognition of Etiology ", " Description of Pathogenesis", "Logical Coherence"],

                "Additional Properties": False,

             },

            " Syndrome Differentiation": {

                "type": "object",

                "properties": {

                    "score": {"type": "number"},

                      "Accuracy of Syndrome": {"type": "number"},

                    "Disease Location and Nature": {"type": "number"}

                },

                "required": ["score", " Accuracy of Syndrome ", "Disease Location and Nature"],

                "Additional Properties": False,

             },

            "Treatment Principle": {

```
            "type": "object",
            "properties": {
                " score ": {"type": "number"},
                "Accuracy of Treatment Principle": {"type": "number"},
                "Specificity of Treatment Method": {"type": "number"},
                "Application of Specialized Methods": {"type": "number"}
            },
            "required": ["score", "Accuracy of Treatment Principle", "Specificity of
Treatment Method", "Application of Specialized Methods"],
            "Additional Properties": False,
        },
        "TCM Prescription": {
            "type": "object",
            "properties": {
                "score": {"type": "number"},
                "Medicinal Match Score": {"type": "number"},
                "Number of matched herbs": {"type": "number"},
                "Number of Herbs in Label Prescription": {"type": "number"},
                "Number of Herbs in Model-Generated Prescription": {"type": "number"},
                "The List of Overlapped Herbs in both TCM Prescriptions ": {"type":
"array", "items": {"type": "string"}},
                "Matching rates": {"type": "string"}
            },
            "required": [
                score", "Medicinal Match Score", "Number of matched herbs": {"type":
"number"}, Number of Herbs in Label Prescription", "Number of Herbs in Model-Generated
Prescription", "The List of Overlapped Herbs in both TCM Prescriptions ", "Matching rates":
{"type": "string"}
            ],
            "Additional Properties": False,
        },
        "Distinguished Theory application": {
            "type": "object",
            "properties": {
                "score": {"type": "number"},
                "Accuracy of Academic Thought": {"type": "number"},
                "Pervasiveness of Thought": {"type": "number"},
                "Completeness of Elaboration": {"type": "number"}
            },
            "required": ["score", "Accuracy of Academic Thought", "Pervasiveness of
Thought", "Completeness of Elaboration"],
            "Additional Properties": False,
        },
        "Total Score": {"type": "number"},
```

```
            "Maximum Score": {"type": "number"}
        },
        "required": [
            "Completeness", "Analysis of Etiology and Pathogenesis", "Syndrome
Differentiation", "Treatment Principle", "TCM Prescription", "Distinguished Theory application",
"Total Score", "Maximum Score"
        ],
        "Additional Properties": False,
    },        "strict": True }
```

**B.2 Expert Evaluation Form for Large Traditional Chinese Medicine Models**

## 1. Your Basic Information

| | | | |
|---|---|---|---|
| Name | _____________ | Sex | _____________ |
| Age | _____ years | Institution Affiliation | / _____________ |
| Professional Title | _____________ | ID Card No. | _____________ |
| Mobile Phone No. | _____________ | | |

## 2. Famous Physician Being Evaluated

Please tick the famous physician you are evaluating:

- Wang, Fengchun
- Qin, Bowei
- Zhao, Shaoqin
- Du, Huaitang
- Gu, Xiaohong

## 3. Rating Items and Evaluation Criteria

This evaluation consists of five real clinical cases. For each case, there are five questions, and the answers of eight models are to be assessed. Each item is scored from 1 to 10; the higher the score, the better the agreement. The estimated time required is 60 minutes.

| Evaluation Dimension | Sub-dimension | Explanation |
|---|---|---|
| **Similarity to distinguished Physicians** | Alignment of aetiology–pathogenesis analysis and pattern differentiation with the famous physician | Degree to which the analysis of aetiology and pathogenesis and the pattern differentiation agree with the famous physician's thinking. |
| | Alignment of therapeutic principles and medication with the famous physician | Degree to which the therapeutic principles and prescribed medicinals agree with the famous physician's thinking. |
| **Consistency with TCM Philosophy** | Consistency of aetiology–pathogenesis analysis and pattern differentiation with TCM theory | Whether the analysis of aetiology and pathogenesis and the pattern differentiation are consistent with traditional Chinese medicine (TCM) theory. |
| | Consistency of therapeutic principles and prescriptions with TCM theory | Whether the therapeutic principles and prescriptions are consistent with TCM theory. |
| **Content Completeness** | | Whether the model's answer to the question is detailed and complete. |
| **Safety** | | Whether the generated prescription is safe and reliable. |
| **Fluency** | | Fluency of the generated text and how easy it is for healthcare professionals to understand. |

## 4. Scoring Table (for Each Case)

Case No.: ________

| Model | Item | score | | | | | | | | | |
|---|---|---|---|---|---|---|---|---|---|---|---|
| **Model1** | **Similarity to distinguished Physicians** | 1 | 2 | 3 | 4 | 5 | 6 | 7 | 8 | 9 | 10 |
| | **Consistency with TCM Philosophy** | 1 | 2 | 3 | 4 | 5 | 6 | 7 | 8 | 9 | 10 |
| | **Content Completeness** | 1 | 2 | 3 | 4 | 5 | 6 | 7 | 8 | 9 | 10 |
| | **Safety** | 1 | 2 | 3 | 4 | 5 | 6 | 7 | 8 | 9 | 10 |
| | **Fluency** | 1 | 2 | 3 | 4 | 5 | 6 | 7 | 8 | 9 | 10 |
| **Model2** | **Similarity to distinguished Physicians** | 1 | 2 | 3 | 4 | 5 | 6 | 7 | 8 | 9 | 10 |
| | **Consistency with TCM Philosophy** | 1 | 2 | 3 | 4 | 5 | 6 | 7 | 8 | 9 | 10 |
| | **Content Completeness** | 1 | 2 | 3 | 4 | 5 | 6 | 7 | 8 | 9 | 10 |
| | **Safety** | 1 | 2 | 3 | 4 | 5 | 6 | 7 | 8 | 9 | 10 |
| | **Fluency** | 1 | 2 | 3 | 4 | 5 | 6 | 7 | 8 | 9 | 10 |
| Model3 | **Similarity to distinguished Physicians** | 1 | 2 | 3 | 4 | 5 | 6 | 7 | 8 | 9 | 10 |
| | **Consistency with TCM Philosophy** | 1 | 2 | 3 | 4 | 5 | 6 | 7 | 8 | 9 | 10 |
| | **Content Completeness** | 1 | 2 | 3 | 4 | 5 | 6 | 7 | 8 | 9 | 10 |
| | **Safety** | 1 | 2 | 3 | 4 | 5 | 6 | 7 | 8 | 9 | 10 |
| | **Fluency** | 1 | 2 | 3 | 4 | 5 | 6 | 7 | 8 | 9 | 10 |
| Model4 | **Similarity to distinguished Physicians** | 1 | 2 | 3 | 4 | 5 | 6 | 7 | 8 | 9 | 10 |
| | **Consistency with TCM Philosophy** | 1 | 2 | 3 | 4 | 5 | 6 | 7 | 8 | 9 | 10 |
| | **Content Completeness** | 1 | 2 | 3 | 4 | 5 | 6 | 7 | 8 | 9 | 10 |
| | **Safety** | 1 | 2 | 3 | 4 | 5 | 6 | 7 | 8 | 9 | 10 |
| | **Fluency** | 1 | 2 | 3 | 4 | 5 | 6 | 7 | 8 | 9 | 10 |
| Model | Item | score | | | | | | | | | |
| Model5 | **Similarity to distinguished Physicians** | 1 | 2 | 3 | 4 | 5 | 6 | 7 | 8 | 9 | 10 |
| | **Consistency with TCM Philosophy** | 1 | 2 | 3 | 4 | 5 | 6 | 7 | 8 | 9 | 10 |
| | **Content Completeness** | 1 | 2 | 3 | 4 | 5 | 6 | 7 | 8 | 9 | 10 |
| | **Safety** | 1 | 2 | 3 | 4 | 5 | 6 | 7 | 8 | 9 | 10 |
| | **Fluency** | 1 | 2 | 3 | 4 | 5 | 6 | 7 | 8 | 9 | 10 |
| Model6 | **Similarity to distinguished Physicians** | 1 | 2 | 3 | 4 | 5 | 6 | 7 | 8 | 9 | 10 |
| | **Consistency with TCM Philosophy** | 1 | 2 | 3 | 4 | 5 | 6 | 7 | 8 | 9 | 10 |
| | **Content Completeness** | 1 | 2 | 3 | 4 | 5 | 6 | 7 | 8 | 9 | 10 |
| | **Safety** | 1 | 2 | 3 | 4 | 5 | 6 | 7 | 8 | 9 | 10 |

| | Fluency | 1 | 2 | 3 | 4 | 5 | 6 | 7 | 8 | 9 | 10 |
|---|---|---|---|---|---|---|---|---|---|---|---|
| **Model7** | **Similarity to distinguished Physicians** | 1 | 2 | 3 | 4 | 5 | 6 | 7 | 8 | 9 | 10 |
| | **Consistency with TCM Philosophy** | 1 | 2 | 3 | 4 | 5 | 6 | 7 | 8 | 9 | 10 |
| | **Content Completeness** | 1 | 2 | 3 | 4 | 5 | 6 | 7 | 8 | 9 | 10 |
| | **Safety** | 1 | 2 | 3 | 4 | 5 | 6 | 7 | 8 | 9 | 10 |
| | **Fluency** | 1 | 2 | 3 | 4 | 5 | 6 | 7 | 8 | 9 | 10 |
| **Model8** | **Similarity to distinguished Physicians** | 1 | 2 | 3 | 4 | 5 | 6 | 7 | 8 | 9 | 10 |
| | **Consistency with TCM Philosophy** | 1 | 2 | 3 | 4 | 5 | 6 | 7 | 8 | 9 | 10 |
| | **Content Completeness** | 1 | 2 | 3 | 4 | 5 | 6 | 7 | 8 | 9 | 10 |
| | **Safety** | 1 | 2 | 3 | 4 | 5 | 6 | 7 | 8 | 9 | 10 |
| | **Fluency** | 1 | 2 | 3 | 4 | 5 | 6 | 7 | 8 | 9 | 10 |

The scores differences between the Med-shicheng and GPT-5, Deepseek-R1, Gemini-2.5-pro, and Qwen3-235B-A22B-Thinking for Dr. Xiaohong Gu, Bowei Qin, Fengchun Wang, and Shaoqin Zhao.

| *Xiaohong Gu* | GPT-5 scores Δ | | | | Deepseek-V3.1 scores Δ | | | |
|---|---|---|---|---|---|---|---|---|
| model | GPT-5 | Deepseek-R1 | Gemini-2.5-pro | Qwen3-235B-A22B-Thinking | GPT-5 | Deepseek-R1 | Gemini-2.5-pro | Qwen3-235B-A22B-Thinking |
| Δ Total | 10.1247 | 6.8332 | 3.0015 | 3.2512 | 6.295 | 5.49 | 0.786 | 2.176 |
| Δ Analysis of etiology and pathogenesis | 1.9 | 1.3 | 1 | 0.4 | 1 | 0.8 | 0.3 | 0.3 |
| Δ Syndrome differentiation | 1.9 | 1.3 | 0.2 | 0.5 | 1 | 0.6 | (0.2) | 0.2 |
| Δ Treatment principle | 4 | 3.4 | 2 | 2.2 | 3 | 2.8 | 2.1 | 2.5 |
| Δ TCM Prescriptions | 0.8247 | (0.0667) | (0.3985) | (0.6487) | 0.195 | 0.29 | (0.414) | (0.324) |
| Δ Distinguished theory application | 1 | 0.4 | (0.3) | 0.3 | 1.9 | 1.2 | 0.6 | 1 |

| *Bowei Qin* | GPT-5 scores Δ | | | | Deepseek-V3.1 scores Δ | | | |
|---|---|---|---|---|---|---|---|---|
| model | GPT-5 | Deepseek-R1 | Gemini-2.5-pro | Qwen3-235B-A22B-Thinking | GPT-5 | Deepseek-R1 | Gemini-2.5-pro | Qwen3-235B-A22B-Thinking |
| Δ Total | 10.368 | 4.6385 | 2.87 | 5.7955 | 7.256 | 6.681 | 5.291 | 6.146 |
| Δ Analysis of etiology and pathogenesis | 2.3 | 1.6 | 0.8 | 1.2 | 2.2 | 1.4 | 1.4 | 1.5 |
| Δ Syndrome differentiation | 1.7 | 0.6 | 0.8 | 0.4 | 2.1 | 1.4 | 1.6 | 1.5 |
| Δ Treatment principle | 3.5 | 1.2 | 0.8 | 2 | 2.8 | 1.9 | 1.8 | 2.1 |
| Δ TCM Prescriptions | 1.068 | 0.4385 | 0.17 | (0.0045) | 0.156 | 0.081 | (0.079) | (0.054) |
| Δ Distinguished theory application | 1.7 | 0.7 | 0.2 | 2.1 | 2.1 | 1 | 0.9 | 1.9 |

| *Fnegchun Wang* | GPT-5 scores Δ | | | | Deepseek-V3.1 scores Δ | | | |
|---|---|---|---|---|---|---|---|---|
| model | GPT-5 | Deepseek-R1 | Gemini-2.5-pro | Qwen3-235B-A22B-Thinking | GPT-5 | Deepseek-R1 | Gemini-2.5-pro | Qwen3-235B-A22B-Thinking |
| Δ Total | 3.5215 | 4.4865 | 1.4275 | 5.825 | (0.95) | 3.645 | (0.315) | 0.5 |

| | | | | | | | |
|---|---|---|---|---|---|---|---|
| Δ Analysis of etiology and pathogenesis | 0.6 | 0.6 | 0.7 | 0.9 | (0.1) | 1.4 | 0.7 | 0.4 |
| Δ Syndrome differentiation | 0.8 | 0.5 | 0.6 | 1.3 | (0.4) | 0.3 | (0.3) | 0.2 |
| Δ Treatment principle | 1.1 | 1.1 | 0.4 | 1.1 | 0.1 | 1.5 | 0.4 | 0.6 |
| Δ TCM Prescriptions | 1.1215 | 0.8865 | 0.0275 | 0.725 | (0.15) | (0.255) | (0.615) | 0 |
| Δ Distinguished theory application | (0.1) | 1.4 | (0.3) | 1.7 | (0.4) | 1.5 | 0.1 | 0.5 |

| *Shaoqin Zhao* | GPT-5 scores Δ | | | | Deepseek-V3.1 scores Δ | | | |
|---|---|---|---|---|---|---|---|---|
| model | GPT-5 | Deepseek-R1 | Gemini-2.5-pro | Qwen3-235B-A22B-Thinking | GPT-5 | Deepseek-R1 | Gemini-2.5-pro | Qwen3-235B-A22B-Thinking |
| Δ Total | 0.3315 | 1.152 | (3.594) | 2.378 | (2.99) | 0.39 | (3.12) | (0.825) |
| Δ Analysis of etiology and pathogenesis | 1 | 1 | 0.1 | 1.4 | (0.2) | 0.5 | (0.4) | 0.2 |
| Δ Syndrome differentiation | 0.7 | 0.8 | 0.4 | 1.2 | (0.6) | 0.7 | (0.3) | 0.3 |
| Δ Treatment principle | (0.6) | (0.6) | (1.8) | (0.4) | (0.9) | 0.7 | (0.4) | 0.2 |
| Δ TCM Prescriptions | 0.1315 | (0.848) | (0.894) | (1.322) | (0.49) | (0.61) | (0.82) | (0.625) |
| Δ Distinguished theory application | (0.9) | 0.8 | (1.4) | 1.5 | (0.9) | 1 | (0.5) | 0.7 |

# D. The responses screenshot of HuatuoGPT2-7B for the clinical trial samples

Fig 1. The responses screenshot of HuatuoGPT2-7B for the clinical trial samples. As we can see that HuatuoGPT2-7B generates the meaningless content given the target clinical trial cases while other comparison LLMs can generate the required response.